\documentclass[journal]{IEEEtran}
\usepackage{amsmath,amsfonts}
\usepackage{rotating}
\usepackage{algorithm}
\usepackage{algpseudocode}
\usepackage{array}
\usepackage[caption=false,font=normalsize,labelfont=sf,textfont=sf]{subfig}
\usepackage{textcomp}
\usepackage{stfloats}
\usepackage{url}
\usepackage{verbatim}
\usepackage{graphicx}
\usepackage{cite}
\usepackage{bm}
\usepackage{amssymb}
\usepackage{graphics}
\usepackage{graphicx}

\usepackage{graphicx}
\usepackage{multirow}
\usepackage{supertabular}

\usepackage{graphicx}
\usepackage[normalem]{ulem}
\useunder{\uline}{\ul}{}

\usepackage{color}
\usepackage{xcolor}
\usepackage{colortbl}
\usepackage{amsmath}
\usepackage{mathtools}
\usepackage{amsthm}
\usepackage{blindtext}
\usepackage{textcomp}
\usepackage{physics}
\usepackage{lscape}
\usepackage{mathrsfs}  
\usepackage{amssymb}
\usepackage{pifont}
\usepackage[english]{babel}

\usepackage{cuted}
\usepackage{etoolbox}
\AfterEndEnvironment{strip}{\leavevmode}
\usepackage[numbers]{natbib}
\begin{document}

\title{Intuitionistic Fuzzy Broad Learning System: Enhancing Robustness Against Noise and Outliers\\}

\author{M. Sajid, A. K. Malik, M. Tanveer{$^*$}, ~\IEEEmembership{Senior Member,~IEEE,} for the Alzheimer’s Disease Neuroimaging Initiative{$^{**}$} 
\thanks{ \noindent $^*$Corresponding Author\\
    M. Sajid, A. K. Malik, and M. Tanveer are with the Department of Mathematics, Indian Institute of Technology Indore, Simrol, Indore, 453552, India (e-mail: phd2101241003@iiti.ac.in, phd1801241003@iiti.ac.in, mtanveer@iiti.ac.in). \\
    \noindent $^{**}$ This study used data from the Alzheimer's Disease Neuroimaging Initiative (ADNI) (adni.loni.usc.edu). The ADNI investigators were responsible for the design and implementation of the dataset, but they did not take part in the analysis or the writing of this publication.  http://adni.loni.usc.edu/wp-content/uploads/how\_to\_apply/ADNI\_Acknowledgement\_List.pdf has a thorough list of ADNI investigators.}}

\markboth{  }%
{Shell \MakeLowercase{\textit{et al.}}: A Sample Article Using IEEEtran.cls for IEEE Journals}


\maketitle

\begin{abstract}
In the realm of data classification, broad learning system (BLS) has proven to be a potent tool that utilizes a layer-by-layer feed-forward neural network. However, the traditional BLS treats all samples as equally significant, which makes it less robust and less effective for real-world datasets with noises and outliers. To address this issue, we propose fuzzy broad learning system (F-BLS) and the intuitionistic fuzzy broad learning system (IF-BLS) models that confront challenges posed by the noise and outliers present in the dataset and enhance overall robustness. Employing a fuzzy membership technique, the proposed F-BLS model embeds sample neighborhood information based on the proximity of each class center within the inherent feature space of the BLS framework. Furthermore, the proposed IF-BLS model introduces intuitionistic fuzzy concepts encompassing membership, non-membership, and score value functions. IF-BLS strategically considers homogeneity and heterogeneity in sample neighborhoods in the kernel space. We evaluate the performance of proposed F-BLS and IF-BLS models on UCI benchmark datasets with and without Gaussian noise. As an application, we implement the proposed F-BLS and IF-BLS models to diagnose Alzheimer's disease (AD). Experimental findings and statistical analyses consistently highlight the superior generalization capabilities of the proposed F-BLS and IF-BLS models over baseline models across all scenarios. The proposed models offer a promising solution to enhance the BLS framework's ability to handle noise and outliers.
\end{abstract} 


\section{Introduction}
\label{introduction}
\IEEEPARstart{T}{he} structure and operations of biological neurons in the brain serve as the basis for the class of machine learning techniques known as artificial neural networks (ANNs). ANNs are composed of interconnected nodes (neurons) that use mathematical operations to process and transfer information. ANNs are designed to discover patterns and correlations in data and then utilize that knowledge to make predictions. Deep learning is one of the most prominent branches of machine learning that employs ANNs with multiple layers to perform complex tasks such as speech recognition \cite{adolfi2023successes}, natural language processing \cite{devlin2018bert}, feature interpretation \cite{2024sajidnfrvfl} and so on.

One of the prime advantages of deep learning is its capacity to automatically extract high-level features from raw data, eliminating the need for manual feature engineering. Deep learning also has the upper hand of being able to generalize effectively to new data, which allows it to make correct predictions even on data that has never been shown to it before. Apart from many advantages, there are some challanges of deep learning architectures: (a) Deep learning models consist of numerous layers and millions of parameters, making them extremely complicated. (b) The iterative learning process in deep learning models can be computationally intensive and time-consuming. (c) Training deep learning models often requires powerful hardware, such as GPUs or TPUs. A shortage of such hardware may hinder the training process. These drawbacks confine the applicability of deep learning models primarily to tasks such as speech and image recognition. Consequently, it gives space to other models to spread their wings.

Randomized neural networks (RNNs) \cite{cao2018review, zhang2016survey} are neural network that incorporates randomness in the topology and learning process of the model. This randomness allows RNNs to learn with fewer tunable parameters in less computational time and without the requirement of savvy hardware. Extreme learning machine (ELM) \cite{huang2006extreme, wang2022review} and random vector functional link neural network (RVFLNN) \cite{pao1994learning, malik2022random} are two popular RNNs. 
Both ELM and RVFLNN use the least-square method (provides a closed-form solution) to determine the output parameters. The direct links from the input layer to the output layer in RVFLNN significantly improve the generalization performance by functioning as an intrinsic regularization tool \cite{zhang2016comprehensive}. In addition, RVFLNN provides quick training speed and universal approximation capability \cite{igelnik1995stochastic}.

 Recently, the broad learning system (BLS) \cite{chen2017broad} (inspired by RVFLNN), a class of flat neural networks, was proposed. BLS employs a layer-by-layer approach to extract informative features from the input data. BLS has three primary segments: a feature learning segment, an enhancement segment, and an output segment. In the feature learning and enhancement segment, the input data is transformed into a high-dimensional feature space using a set of random projection mappings.
 As a result of the feature learning and enhancement segment, a collection of high-dimensional features is produced that effectively extract important information from the input data. 
 To train the BLS model, only the output layer parameters need to be computed by the least-squares method. In comparison to traditional deep neural networks (DNNs), BLS has several advantages. For example, it does not employ backpropagation, which might simplify the training process and lessen the likelihood of overfitting; its ability to learn from a small number of training samples; and faster training speed due to closed form solution. The universal approximation capability of BLS \cite{chen2018universal} makes it more promising among researchers. The adaptable topology of BLS makes it possible to train and update the model effectively in an incremental way \cite{chen2017broad}, and when training data is scarce, BLS may have superior generalization performance than deep learning models \cite{yu2019broad}.

There has been a range of distinct variants of the BLS architecture proposed in the literature, each with its unique features, advantages, and limitations.  The double-kernelized weighted BLS (DKWBLS) \cite{chen2022double} was developed to manage the imbalanced data.  \citet{wang2021multi} proposed a BLS-based multi-modal material identification paradigm. This technique lowers the joint feature dimensionality by optimizing the correlation between two features of the input samples. 
In \cite{feng2018broad}, the authors proposed a time-varying iterative learning algorithm utilizing gradient descent (GD) for optimizing the parameters in the enhancement layer, feature layer, and output layer. In \cite{feng2018fuzzy}, a hybrid neuro-fuzzy BLS (NeuroFBLS) was proposed by combining a human-like reasoning approach based on a set of IF-THEN fuzzy rules with the learning and linking structure of the BLS. NeuroFBLS employs the k-means clustering technique to create fuzzy subsystems, thereby increasing the computational burden of the model.

BLS has a number of benefits over conventional machine learning and deep learning models \cite{gong2021research}.
However, BLS is susceptible to noise and outliers.  Outliers are data points that deviate greatly from the majority of the data points, whereas noise refers to random abnormalities or oscillations in the data \cite{smiti2020critical}. The factors contributing to BLS's susceptibility to noise and outliers are as follows.
(i) When the noisy features of a sample propagate through the feature and enhancement groups of the BLS, then all features (pure and impure) get mixed with each other. 
As a result, noisy features corrupt the higher-level features of enhancement groups, consequently skewing the learning process 
of the BLS. (ii) In the BLS model, every sample is assigned a uniform weight during training, regardless of the sample's purity or abnormality. Ideally, noisy samples should be given less weight than pure ones. However, BLS overlooks this distinction, resulting in difficulty discerning between pure and impure samples and consequently yielding poor generalization performance in such scenarios.

Fuzzy theory is successfully employed in machine learning models \cite{lin2002fuzzy, Malik2022, ganaie2023graph} to ameliorate the adverse effects of noise or outliers on model performance. The fuzzy membership function uses sample-to-class center distance to generate a degree of membership to deal with outliers and noise. In \cite{ha2013support}, intuitionistic fuzzy (IF) score was presented as an advanced version of the fuzzy membership scheme. The IF theory uses the membership and non-membership functions to assign an IF score to each sample to deal with the noise and outliers efficiently. Membership and non-membership values of a sample measure the degree of belongingness and non-belongingness of that sample to a particular class, respectively.

To get motivated by the remarkable ability of fuzzy theory to handle the noise and outliers in the data, in this paper, we amalgamate the fuzzy and intuitionistic fuzzy theory with the BLS model. We propose two novel models, namely fuzzy BLS (F-BLS) and intuitionistic fuzzy BLS (IF-BLS) to cope with the noisy samples and the outliers that have trespassed in the dataset. 
The following are the main highlights of this paper:
\begin{enumerate}
    \item We propose a novel fuzzy BLS (F-BLS), wherein fuzzy membership assigns a different weightage to each sample based on its proximity to the class center. This adaptive weighting mechanism enables distinct treatment of noise and outlier samples compared to pure ones.
    \item Furthermore, we propose intuitionistic fuzzy BLS (IF-BLS) model, which incorporates IF concepts involving membership, non-membership, and score functions. IF-BLS takes into account homogeneity and heterogeneity within sample neighborhoods in the kernel space, facilitating the precise assignment of weights based on each sample's neighborhood information and proximity.
    \item Experimental evaluations on UCI benchmark datasets, both with and without Gaussian noise, across diverse domains demonstrate the superior robustness of our proposed F-BLS and IF-BLS models compared to baseline models in situations where noise and outliers pose significant challenges.
    \item The application of the proposed F-BLS and IF-BLS models in Alzheimer's disease (AD) diagnosis showcases their superior performance over baseline models.
\end{enumerate}
The remaining structure of the paper is organized as follows. In Section \ref{Broad Learning system}, we provide a brief overview of BLS, fuzzy and IF schemes. In Section \ref{proposed_work}, we derive the mathematical formulation of the proposed F-BLS and IF-BLS models. In Section \ref{experiments}, we discuss experimental findings across diverse scenarios and datasets. Finally, we conclude the paper in Section \ref{conclusion} by suggesting future research directions.
\section{Related Works}
\label{Broad Learning system}
In this section, we go through the architecture of BLS along with its mathematical formulation, fuzzy membership schemes, and intuitionistic fuzzy membership.
Let $\bigl\{\left(X,T\right) \vert\, X \in \mathbb{R}^{N \times D},\, T \in \mathbb{R}^{N \times C}\big\}$ be the training set, where $X$ is the input matrix and $T$ is the target matrix. Here, $N$ denotes the total number of training samples, each of which has dimension $D$ and $C$ number of classes. 
\subsection{Broad Learning System (BLS) \cite{chen2017broad}}
\label{bls_subsection}
\begin{figure}[h]
\centering
\includegraphics[width=0.5\textwidth]{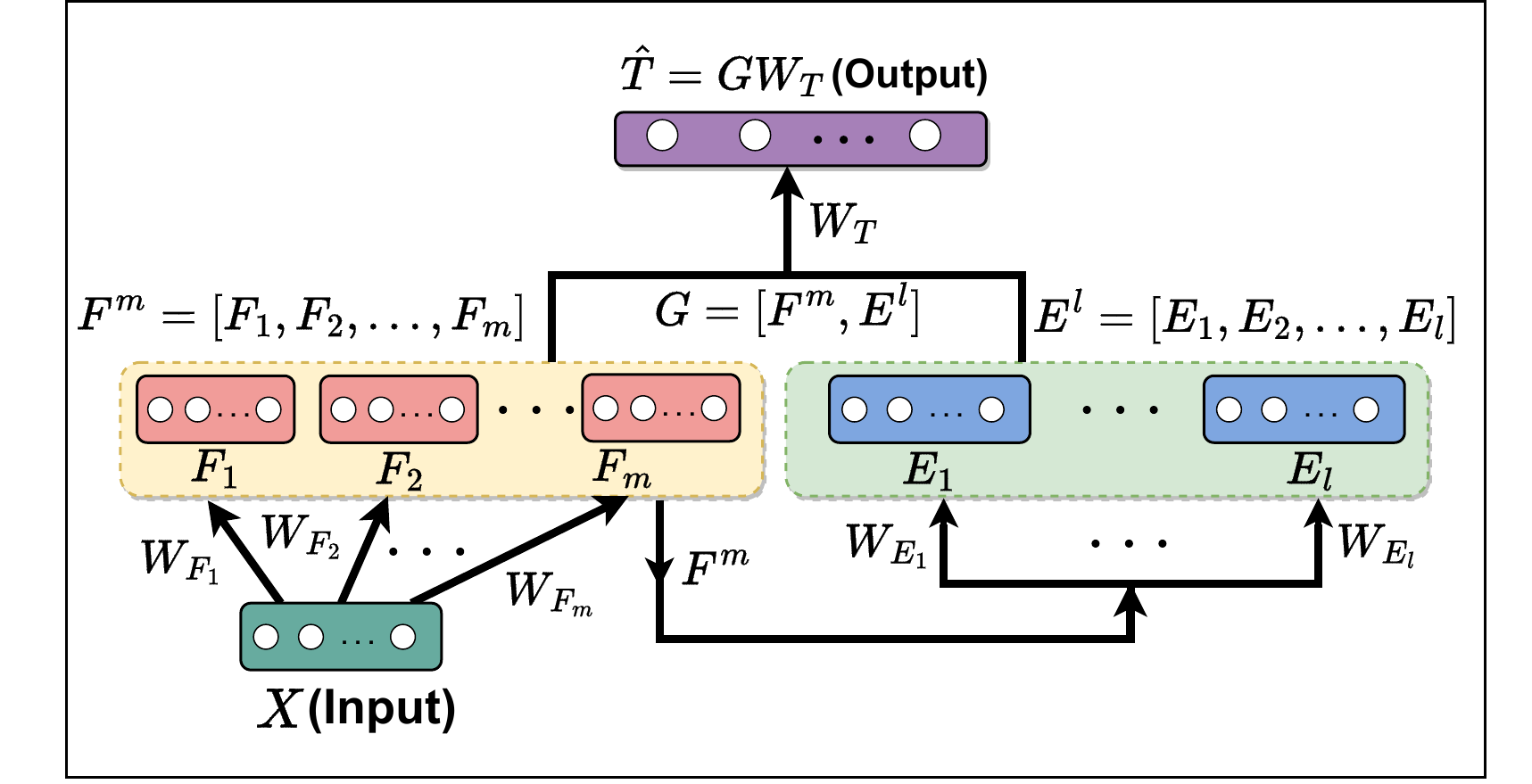}
\caption{The architecture of the BLS model.}
\label{diag:BLS}
\end{figure}
The architecture of the BLS is shown in Figure \ref{diag:BLS}. We briefly give the mathematical formulations of the BLS model.\\
\textbf{Segment-1:} Let there be $m$ groups and each feature group having $p$ nodes. Then the $i^{th}$ feature group is given as follows:
\begin{equation}
    \label{eq:F_i}
F_i=\mathcal{F}_i\left(XW_{F_{i}}+\beta_{F_{i}}\right) \in \mathbb{R}^{N \times p}, ~~ i=1,2,\hdots,m,
\end{equation}
where $\mathcal{F}i$, $W_{F_{i}} \in \mathbb{R}^{D \times p}$ and $\beta_{F_{i}} \in \mathbb{R}^{N \times p}$ are feature map, randomly generated weight matrix, and bias matrix for the $i^{th}$ feature group, respectively. The augmented output of the $m$ number of feature groups is: 
\begin{equation}
\label{eq:F^m}
    F^m=[F_1,F_2,\hdots,F_m] \in \mathbb{R}^{N \times mp}.
\end{equation}
\textbf{Segment-2:} Augmented feature matrix $(F^m)$ is projected to enhancement spaces via random transformations followed by the activation functions. Let $l$ be the number of enhancement groups, and each enhancement group has $q$ nodes. Then
\begin{equation}
    \label{eq:E_j}
E_j=\chi_j\left(F^mW_{E_{j}}+\beta_{E_{j}}\right) \in \mathbb{R}^{N \times q}, ~~ j=1,2,\hdots,l.
\end{equation}
Where $\chi_j$ is the activation function used to generate nonlinear features. $W_{E_{j}} \in \mathbb{R}^{mp \times q}$ is a randomly generated weight matrix connecting the augmented feature matrix $(F^m)$ to $j^{th}$ enhancement group, and $\beta_{E_{j}} \in \mathbb{R}^{N \times q}$ is the bias matrix corresponding to the $j^{th}$ enhancement group. The resultant output of the $l$ enhancement groups is 
\begin{equation}
\label{eq:E^l}
    E^l=[E_1,E_2,\hdots,E_l] \in \mathbb{R}^{N \times lq}.
\end{equation}
\textbf{Segment-3:} Finally, the resultant enhancement matrix $(E^l)$ along with the augmented feature matrix $(F^m)$ is sent to the output layer and the final outcome is calculated as follows: 
\begin{align}
    \Hat{T}=[F^m,E^l]W_T =GW_T,
\end{align}
where 
\begin{align}
\label{eq:G}
G=[F^m,E^l]  \in \mathbb{R}^{N \times (mp+lq)}
\end{align}
is the concatenated matrix, and $W_T \in \mathbb{R}^{(mp+lq) \times C}$ 
is the unknown weights matrix connecting the augmented feature layer and resultant enhancement layer to the output layer. Finally, $W_T$ is calculated using the pseudo-inverse as: $W_T=G^{\dagger}\Hat{T}$,
where $G^{\dagger}$ denotes the pseudo-inverse of $G$.
\subsection{Fuzzy Membership Scheme \cite{Lin2002}}
\label{Fuzzy Membership Scheme}
Let $X$ be the crisp set and $\vartheta: X \rightarrow [0,1]$ be the membership function which assigns fuzzy membership values to each sample $x_r \in X$. Define $\mathcal{A}=\bigl\{(x_r,t_r,\vartheta(x_r))\vert\, x_r \in \mathbb{R}^{1 \times D},\, t_r \in \mathbb{R}^{1 \times C},\,  r=1,2,\hdots,N\bigl\}$ be the fuzzy set. 
The membership function for training sample $x_r$ is defined as:
\begin{equation}
\label{eq:mu(x_r)}
\vartheta\left(x_r\right)=\left\{\begin{array}{ll}
1-\frac{\left\|x_r-C_{pos}\right\|}{R_{pos}+\delta}, & t_r=+1, \\
1-\frac{\left\|x_r-C_{neg}\right\|}{R_{neg}+\delta}, & t_r=-1,
\end{array}\right.
\end{equation}
where $C_{pos} (C_{neg})$ and $R_{pos} (R_{neg})$ are the class center and the
the radius of the positive (negative) class, respectively, and $\delta$ is a very small positive parameter. $t_r=+1 ~(-1)$ denotes the positive (negative) class. The membership function takes into account the distance between each sample and its associated class center. The center of each class is defined as: 
\begin{equation}
\label{centre}
    C_{pos}=\frac{1}{N_{pos}}\sum_{t_r=+1}x_r ~~\text{and}~~ C_{neg}=\frac{1}{N_{neg}}\sum_{t_r=-1}x_r,
\end{equation}
where $N_{pos}$ ($N_{neg}$) is the total number of samples in the positive (negative) class. The radii are defined as:
\begin{equation}
\label{radius}
    R_{pos}=\underset{{t_r=+1}}\max\|x_r - C_{pos}\| ~~\text{and}~~ R_{neg}=\underset{{t_r=-1}}\max\|x_r - C_{neg}\|.
\end{equation}
Finally fuzzy score matrix $\mathcal{S}$ for the dataset $X$ is defined as:
\begin{align}
\label{eq:FBLS_Score_matrix}
\mathcal{S}=\text{$diag$}\left(\vartheta(x_1), \vartheta(x_2), \hdots, \vartheta(x_N)\right).
\end{align}
\subsection{Intuitionistic Fuzzy Membership (IFM) Scheme \cite{ha2013support}}
\label{Intuitionistic Fuzzy Membership (IFM) Scheme}
The intuitionistic fuzzy set (IFS) \cite{atanassov1999intuitionistic} was defined to cope with uncertainty difficulties, and it allows for a precise simulation of the situation using current information and observations \cite{ha2013support}. Here, the IFM scheme for the crisp set $X$ is discussed by projecting the dataset $X$ onto a kernel space. 
An IFS is defined as: $ \widetilde{\mathcal{A}}=\bigl\{(x_r,t_r,\theta(x_r), \widetilde\theta(x_r)\vert\,\, r=1,2,\hdots,N\bigl\},$
where $\theta(x_r)$ and $\widetilde\theta(x_r)$ are the membership and non-membership values of $x_r$.
\begin{itemize}
\item{\bf{The membership mapping:}} The membership function $\theta: X \rightarrow [0,1]$ is defined as: 
\begin{equation}
\label{membership}
\theta\left(x_r\right)=\left\{\begin{array}{ll}
1-\frac{\left\|\psi(x_r)-C_{pos}\right\|}{R_{pos}+\delta}, & t_r=+1, \\
1-\frac{\left\|\psi(x_r)-C_{neg}\right\|}{R_{neg}+\delta}, & t_r=-1,
\end{array}\right.
\end{equation}
where $\psi$ represents the kernel mapping. $C_{pos}$, $C_{neg}$, $R_{pos}$ \text{and} $R_{neg}$ are defined in the same way as mentioned in Subsection \ref{Fuzzy Membership Scheme}, with replacing $x_r$ by $\psi(x_r)$.
\item{\bf{The non-membership mapping:}} In the IFM scheme, each training sample $x_r$ is additionally allocated a non-membership value that represents the ratio of heterogeneous data points to all data points in its vicinity. The non-membership function $\widetilde\theta: X \rightarrow [0,1]$ is defined as: $\widetilde\theta(x_r)=\big(1-\theta(x_r)\big)\Theta(x_r),$
where $0\leq \theta(x_r)+\widetilde\theta(x_r)\leq 1$ and 
$\Theta$ is calculated as follows:
\begin{equation}
\label{eq:nonmembership}
    \Theta(x_r)=\frac{\vert\{x_l:\|\psi(x_r)-\psi(x_l)\|\leq\epsilon,~y_r\neq y_l\}\vert}{\vert\{x_l:\|\psi(x_r)-\psi(x_l)\|\leq\epsilon\}\vert},
\end{equation}
where $\epsilon$ is the adjustable parameter used to create a neighborhood and $\vert \cdot \vert$ represents the cardinality of a set.
\item{\bf{The score mapping:}} After calculating the membership and non-membership values of each sample, an IF score (IFS) function is defined as follows:
\begin{equation}
\label{eq:score_function}
    \Theta^{\star}(x_r)=\left\{\begin{array}{lll}{\theta(x_r)}, & {\widetilde\theta(x_r)=0}, \\ 
    0, & \theta(x_r) \leq \widetilde\theta(x_r), \\ \frac{1-\widetilde\theta(x_r)}{2-\theta(x_r)-\widetilde\theta(x_r)}, & \text{otherwise}.\end{array}\right. 
\end{equation}
\item{\bf{The score matrix:}} Finally, the score matrix $\mathcal{S}$ for the dataset $X$ is defined as: 
\begin{align}
\label{eq:IFBLS_Score_matrix}
\mathcal{S}=\text{$diag$}\left(\Theta^{\star}(x_1), \Theta^{\star}(x_2), \hdots, \Theta^{\star}(x_N)\right).
\end{align}
\end{itemize}
The kernel technique is explored in supplementary Section S.I.
\section{Proposed Fuzzy Broad Learning System (F-BLS) and Intuitionistic Fuzzy Broad Learning System (IF-BLS)}
\label{proposed_work}
In BLS, each data sample is given the same weight irrespective of its nature. Naturally, a dataset is never pure, $i.e.,$ the involvement of noise and outliers in a dataset is a normal phenomenon. 
Although the occurrence of these noises and outliers is natural, it has a detrimental impact on the traditional BLS. Therefore, to deal with noisy samples and outliers present in the dataset, we propose two models:
\begin{enumerate}
    \item Fuzzy BLS (F-BLS) by employing the fuzzy membership scheme discussed in the subsection \ref{Fuzzy Membership Scheme}.
    \item Intuitionistic fuzzy BLS (IF-BLS) using the IF scheme discussed in \ref{Intuitionistic Fuzzy Membership (IFM) Scheme}. 
\end{enumerate}
The fuzzy membership value in the proposed F-BLS and IF-BLS models is based on how close a sample is to the corresponding class center in the original feature space and kernel space, respectively. These fuzzy membership value represent the degree to which a sample belongs to a certain class. In IF-BLS, the fuzzy non-membership value is determined by considering the sample's neighborhood information, indicating its extent of non-belongingness to a specific class.

\begin{figure}[h]
\centering
\includegraphics[width=0.45\textwidth]{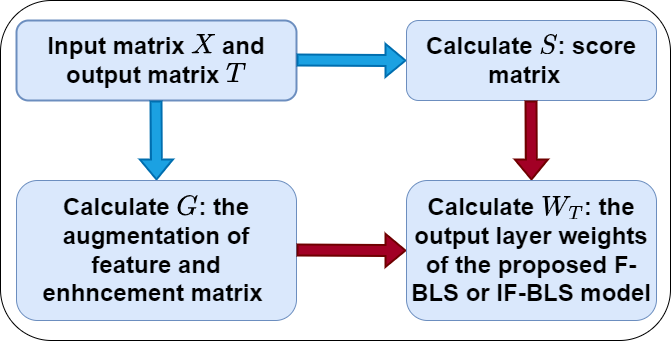}
\caption{The flowchart of finding the output weight matrix $W_T$ of the proposed F-BLS and IF-BLS models.}
\label{flowchart:IFBLS}
\end{figure}

The optimization problem of the proposed F-BLS and IF-BLS models are defined as follows:  
\begin{align}   
\label{eq20}
&W_{T_{\min}}=\underset{{W_T}}{\text{argmin}}\,  \frac{C}{2}\|\mathcal{S}\zeta\|_2^2+\frac{1}{2}\|W_T\|_2^2, \nonumber \\
& \text{s.t.} ~~G W_T-T=\zeta,
\end{align} 
where $\mathcal{S}$ denotes the score matrix. For F-BLS, the score matrix is obtained from the Subsection \ref{Fuzzy Membership Scheme}, whereas for IF-BLS, $\mathcal{S}$ is taken from the Subsection \ref{Intuitionistic Fuzzy Membership (IFM) Scheme}. Here, $\zeta$ refers to the error matrix. Equivalently, the formulation \eqref{eq20} can be written as:
\begin{align}  
\label{eq21}
W_{T_{\min}}=\underset{{W_T}}{\text{argmin}}\,  \frac{C}{2}\|\mathcal{S} ( G W_T-T)\|_2^2+\frac{1}{2}\|W_T\|_2^2.
\end{align}
Problem \eqref{eq21} is the convex quadratic programming problem (QPP) and hence possesses a unique solution. The Lagrangian of \eqref{eq20} is written as:
\begin{align}   
\label{eq22}
\mathcal{L}(W_T,\zeta,\lambda)=\frac{C}{2}\|\mathcal{S}\zeta\|_2^2+\frac{1}{2}\|W_T\|_2^2-\lambda^t(G W_T-T-\zeta),
\end{align}
where $\lambda$ is the Lagrangian multiplier and $(.)^t$ is the transpose operator. Differentiating $\mathcal{L}$ partially  w.r.t. each parameters, $i.e., W_T,\zeta ~\text{and}~ \lambda$; and equating them to zero, we obtain
\begin{align}   
& \frac{\partial \mathcal{L}}{\partial W_T}=0 \Rightarrow W_T - G^t \lambda = 0 \Rightarrow W_T = G^t \lambda, \label{eq23}\\
& \frac{\partial \mathcal{L}}{\partial \zeta}=0 \Rightarrow C \mathcal{S}^t (\mathcal{S} \zeta) +\lambda = 0 \Rightarrow \lambda = - C \mathcal{S}^t \mathcal{S} \zeta,\label{eq24}\\
& \frac{\partial \mathcal{L}}{\partial \lambda}=0 \Rightarrow G W_T-T-\zeta = 0 \Rightarrow \zeta = G W_T-T. \label{eq25}
\end{align}
Substituting Eq. \eqref{eq25} in \eqref{eq24}, we get
\begin{align}  
\label{eq26}
\lambda = - C \mathcal{S}^t \mathcal{S} (G W_T-T).
\end{align}
On substituting the value of $\lambda$ obtained in Eq. \eqref{eq23}, we obtain
\begin{align} 
&W_T = G^t\left(- C \mathcal{S}^t \mathcal{S} (G W_T-T)\right), \label{eq27a} \\
& \Rightarrow W_T = -C G^t\mathcal{S}^2 G W_T + C G^t\mathcal{S}^2 T, \label{eq27b}\\
& \Rightarrow \left(I + C G^t\mathcal{S}^2 G \right) W_T = C G^t\mathcal{S}^2 T, \label{eq27c}\\
& \Rightarrow W_T = \left(G^t\mathcal{S}^2G + \frac{1}{C} I\right)^{-1} G^t \mathcal{S}^2T, \label{eq27d}
\end{align}
where $I$ is the identity matrix of the appropriate dimension. Substituting the values of \eqref{eq25} and \eqref{eq23} in \eqref{eq24}, we get
\begin{align} 
\label{eq29a}
&\lambda = -{C}\mathcal{S}^t\mathcal{S} (GG^t\lambda -T),\\
&\Rightarrow \lambda + {C}\mathcal{S}^2 GG^t\lambda = C\mathcal{S}^2T, \label{eq29b}\\
&\Rightarrow C\left(\frac{1}{C}I + \mathcal{S}^2 GG^t\right)\lambda = C\mathcal{S}^2T, \label{eq29c}\\
&\Rightarrow \lambda = \left(\frac{1}{C}I + \mathcal{S}^2 GG^t\right)^{-1}\mathcal{S}^2T.\label{eq29d}
\end{align}
Putting the value of  $\lambda$ obtained in Eq. \eqref{eq29d} in \eqref{eq23}, we get
\begin{align} 
W_T = G^t
\left(\frac{1}{C}I+\mathcal{S}^2GG^t\right)^{-1}\mathcal{S}^2T. \label{eq31}
\end{align}
We get two distinct formulas, \eqref{eq27d} and \eqref{eq31}, that can be utilized to determine $W_T$. It is worth noting that both formulas involve the calculation of the matrix inverse. 
If the number of features ($mp+lq$) in $G$ is less than or equal to the number of samples ($N$), we employ the formula \eqref{eq27d} to compute $W_T$. Otherwise, we opt for the formula \eqref{eq31} to calculate $W_T$. As a result, we possess the advantage of calculating the matrix inverse either in the feature or sample space, contingent upon the specific scenario. Therefore, the optimal solution of \eqref{eq20} is given as:
\begin{equation}
\label{eq:W_T_IFBLS}
    W_T=\left\{\begin{array}{ll}{\left(G^t\mathcal{S}^2G + \frac{1}{C} I \right)^{-1} G^t \mathcal{S}^2T}, & \text{if}~~ (mp+lq) \leq N,  \vspace{3mm} \\ {G^t
\left(\frac{1}{C} I+\mathcal{S}^2GG^t\right)^{-1}\mathcal{S}^2T} , & \text{if}~~ N < (mp+lq).\end{array}\right. 
\end{equation}
\textbf{\textit{Remarks:}} The Gaussian kernel function is employed to project the input samples into a higher-dimensional space. The Gaussian kernel is defined as: $K(x_1,x_2)=exp(-\frac{\norm{x_1-x_2}^2}{\mu^2})$, where $\mu$ is the kernel parameter.\\
The flowchart presented in Figure \ref{flowchart:IFBLS} illustrates the operational steps outlined for the Algorithm \ref{alg:alogorithm1} of the proposed models.
\subsection{Complexity Analysis of the Proposed F-BLS and IF-BLS}  
Consider a dataset consisting of $N$ samples. In IF-BLS, both membership and non-membership values are computed for each individual sample. As a result, the computational effort required to calculate these values is of $\mathcal{O}(N)$ \cite{rezvani2019intuitionistic}. The complexity of the proposed IF-BLS primarily depends on the matrix inversion computation. Following the standard procedure \cite{zhang2015divide}, the complexity of inverting a matrix of size $N\times N$ is $\mathcal{O}(N^3)$. For large values of $N$, $\mathcal{O}(N)$ becomes negligible in comparison to the $\mathcal{O}(N^3)$. Consequently, considering the complexity of the entire IF-BLS model, it can be approximated as $\mathcal{O}(N^3)$ due to the dominant contribution of the matrix inversion term. Since, in F-BLS, only membership values need to be calculated, therefore, the computational complexity of fuzzy membership is also $\mathcal{O}(N)$. Similarly, the overall computational complexity of F-BLS is also $\mathcal{O}(N^3)$.
\begin{algorithm}
\caption{Algorithm of the proposed F-BLS and IF-BLS models.}\label{alg:alogorithm1}
\begin{algorithmic}[1]
\State \textbf{Input:} Dataset of $N$ sample, $D$ features and $C$ classes.
\State \textbf{Input:} Input matrix $X \in \mathbb{R}^{N \times D}$ and output matrix $T \in \mathbb{R}^{N \times C}$.
\State \textbf{Parameters:} $\mathcal{C}, \mu, m, p, l ~\text{and}~ q$ represents regularization parameter, kernel parameter, \#feature groups, \#feature nodes, \#enhancement groups  and \#enhancement nodes, respectively. 
\State \textbf{Input Sample:} Let $x_r$ be the $r^{th}$ training sample, where $r=1,2,\hdots, N$. 
\State \textbf{Find}: $F_i \in \mathbb{R}^{N \times p}, ~~ i=1,2,\hdots,m,$ using Eq. \eqref{eq:F_i}.
\State \textbf{Find}: $F^m \in \mathbb{R}^{N \times mp}$ using Eq. \eqref{eq:F^m}.
\State \textbf{Find}: $E_j \in \mathbb{R}^{N \times q}, ~~ j=1,2,\hdots,l,$ using Eq. \eqref{eq:E_j}.
\State \textbf{Find}: $E^l \in \mathbb{R}^{N \times lq}$ using Eq. \eqref{eq:E^l}.
\State \textbf{Find}: $G \in \mathbb{R}^{N \times (mp+lq)}$ using Eq. \eqref{eq:G}.

\vspace{-3mm} \hrulefill \\
\textbf{If Model$=$F-BLS, follow steps (11) - (12), else Model$=$IF-BLS, follow steps (13) - (16).}

\vspace{-3mm} \hrulefill
\State \textbf{Find}: Membership value $\vartheta\left(x_r\right)$ for $x_r$ using Eq. \eqref{eq:mu(x_r)}.
\State \textbf{Find}: Score matrix $\mathcal{S}$ using Eq. \eqref{eq:FBLS_Score_matrix}.

\vspace{-3mm} \hrulefill

\State \textbf{Find}: Membership value $\theta\left(x_r\right)$ for $x_r$ using Eq. \eqref{membership}.
\State \textbf{Find}: Non-membership value $\Theta\left(x_r\right)$ for $x_r$ using Eq. \eqref{eq:nonmembership}.
\State \textbf{Find}: Intutionistic fuzzy score value $\Theta^{\star}(x_r)$ using \eqref{eq:score_function}.
\State \textbf{Find}: Score matrix $\mathcal{S}$ using Eq. \eqref{eq:IFBLS_Score_matrix}.

\vspace{-3mm} \hrulefill

\State \textbf{Output}: Output weight matrix $W_T$ using Eq. \eqref{eq:W_T_IFBLS}.
\end{algorithmic}
\vspace{-3mm} \hrulefill \\
Source code link: \emph{https://github.com/mtanveer1/IF-BLS}.
\end{algorithm}
\begin{table*}[]
\centering
\caption{The testing accuracy of the proposed F-BLS and IF-BLS models along with the baseline models on UCI datasets.}
\label{tab:UCI_Testing_Accuracy}
\resizebox{16.5cm}{!}{%
\begin{tabular}{lccccccccc}  \hline \vspace{-3mm}\\ 
\textbf{Dataset $\downarrow$ $\mid$ \text{Model} $\rightarrow$} &
  \textbf{\#Samples} &
  \textbf{\#Features} &
  \textbf{BLS \cite{chen2017broad}} &
  \textbf{ELM \cite{huang2006extreme}} &
  \textbf{NeuroFBLS \cite{feng2018fuzzy}} &
  \textbf{IF-TSVM \cite{rezvani2019intuitionistic}} &
  \textbf{H-ELM \cite{tang2015extreme}} &
  \textbf{F-BLS $^{\dagger}$} &
  \textbf{IF-BLS $^{\dagger}$} \vspace{0mm}\\ \hline \vspace{-3mm}\\
acute\_inflammation               & 120  & 7   & \textbf{100}     & \textbf{100}     & \textbf{100}     & 40.8333 & \textbf{100}     & \textbf{100}     & 100     \\
acute\_nephritis                  & 120  & 7   & \textbf{100}     & \textbf{100}     & \textbf{100}     & 51.6667 & 82.5    & \textbf{100}     & \textbf{100}     \\
bank                              & 4521 & 17  & 89.7366 & 89.4051 & 89.5817 & 88.4981 & 88.6308 & \textbf{89.759}  & 89.4051 \\
breast\_cancer                    & 286  & 10  & 69.8851 & 66.6727 & 70.1754 & 70.1754 & \textbf{84.271}  & 72.2868 & 83.1579 \\
breast\_cancer\_wisc              & 699  & 10  & 88.4183 & 87.9897 & \textbf{90.7081} & 66.6783 & 85.4162 & 88.2775 & 88.9938 \\
chess\_krvkp                      & 3196 & 37  & 84.3862 & 72.0312 & 70.4004 & 25.1017 & 82.1028 & 84.2921 & \textbf{84.9184} \\
conn\_bench\_sonar\_mines\_rocks  & 208  & 61  & 69.2451 & 60.5807 & 60.6272 & 22.7758 & 75.4936 & 69.1521 & \textbf{80.2091} \\
credit\_approval                  & 690  & 16  & 87.5362 & 85.3623 & 84.4928 & 37.5362 & 86.8116 & 86.3768 & \textbf{88.5507} \\
cylinder\_bands                   & 512  & 36  & 69.9258 & 66.2193 & 69.5336 & 60.8719 & 63.242  & 69.1338 & \textbf{72.8536} \\
echocardiogram                    & 131  & 11  & 83.9316 & 83.9031 & 80.9402 & 80.0855 & 84.6724 & 84.6724 & \textbf{88.49}   \\
fertility                         & 100  & 10  & 90      & 89      & \textbf{92}     & 88      & 89      & 91      & 91      \\
haberman\_survival                & 306  & 4   & 70.275  & 73.4902 & 73.4902 & 73.4902 & 74.146  & 69.6563 & \textbf{75.4574} \\
hepatitis                         & 155  & 20  & 85.1613 & 83.2258 & \textbf{87.7419} & 79.3548 & 82.5806 & 84.5161 & \textbf{87.7419} \\
hill\_valley                      & 1212 & 101 & \textbf{82.0185} & 77.9706 & 78.9583 & 53.717  & 71.6155 & 81.6002 & 79.6249 \\
horse\_colic                      & 368  & 26  & 86.1422 & 84.7908 & 83.9726 & 63.0285 & 86.1459 & 86.1385 & \textbf{86.6938} \\
mammographic                      & 961  & 6   & 78.6712 & 79.0889 & 79.504  & 53.689  & 72.3208 & 78.6744 & \textbf{79.8192} \\
molec\_biol\_promoter             & 106  & 58  & 83.9394 & 68.961  & 65.1515 & 25.7143 & 82.1212 & 84.9351 & \textbf{88.7879} \\
monks\_1                          & 556  & 7   & 75.3314 & 83.2497 & \textbf{86.1277} & 40.6403 & 66.7165 & 77.1396 & 77.6705 \\
musk\_1                           & 476  & 167 & 75.8311 & 67.864  & 67.8487 & 22.1053 & \textbf{89.4868} & 76.6754 & 78.9846 \\
oocytes\_merluccius\_nucleus\_4d  & 1022 & 42  & \textbf{82.7776} & 79.8407 & 80.142  & 67.219  & 70.6428 & 81.8011 & 80.9201 \\
oocytes\_trisopterus\_nucleus\_2f & 912  & 26  & \textbf{78.9371} & 76.4211 & 75.9911 & 58.1229 & 71.3799 & 78.6123 & 75.9899 \\
pima                              & 768  & 9   & 72.0058 & 71.4888 & 72.7918 & 65.1006 & 70.3081 & 71.4846 & \textbf{73.3079} \\
pittsburg\_bridges\_T\_OR\_D      & 102  & 8   & 89.1905 & 88.1429 & 90.1429 & 86.1429 & 87.2857 & 88.1905 & \textbf{90.2381} \\
spambase                          & 4601 & 58  & 89.0021 & 87.1125 & 83.2657 & 60.6087 & 90.5259 & 89.4582 & \textbf{90.7407} \\
spect                             & 265  & 23  & 69.434  & 67.9245 & 67.9245 & 58.4906 & 66.7925 & 69.434  & \textbf{72.4528} \\
statlog\_heart                    & 270  & 14  & 82.2222 & 80      & 80.7407 & 55.5556 & 80.7407 & 82.2222 & \textbf{84.0741}\\
tic\_tac\_toe                     & 958  & 10  & \textbf{98.4315} & 86.0068 & 82.7623 & 65.3125 & 75.7292 & 97.8081 & 97.0768 \\
titanic                           & 2201 & 4   & 77.9168 & 77.9168 & 78.0537 & 73.9173 & 77.3264 & 77.9168 & \textbf{79.0532} \vspace{0mm}\\ \hline \vspace{-3mm}\\
\multicolumn{3}{c}{Average Accuracy}           & 82.5126 & 79.8093 & 80.1096 & 58.3726 & 79.9287 & \underline{82.5434} & \textbf{84.5076} \vspace{0mm}\\ \hline \vspace{-3mm}\\
\multicolumn{3}{c}{Average Standard Deviation} & 6.5829  & 6.7853  & 7.4441  & 17.6829 & 10.3037 & \underline{6.1989}  & \textbf{6.0494}  \vspace{0mm}\\ \hline \vspace{-3mm}\\
\multicolumn{3}{c}{Average Rank}               & \underline{3.0893}  & 4.6071  & 3.8036  & 6.8036  & 4.5357  & 3.2679  & \textbf{1.8929} \vspace{0mm}\\ \hline \vspace{-3mm}\\
\multicolumn{10}{l}{The boldface and underline in the row denote the best and second-best performed model corresponding to the datasets. $\dagger$ represents the proposed models.}
\end{tabular}%
}
\end{table*}
\begin{table*}[]
\centering
\caption{Wilcoxon signed-rank test of the proposed F-BLS and IF-BLS models \textit{w.r.t.} baseline models on UCI datasets.}
\label{tab:UCI_Wilcoxon}
\resizebox{17cm}{!}{%
\begin{tabular}{|l|cc|cc|cc|cc|cc|}
\hline
\textbf{\text{Baseline} $\rightarrow$} &
  \multicolumn{2}{c|}{\textbf{BLS}} &
  \multicolumn{2}{c|}{\textbf{ELM}} &
  \multicolumn{2}{c|}{\textbf{NeuroFBLS}} &
  \multicolumn{2}{c|}{\textbf{IF-TSVM}} &
  \multicolumn{2}{c|}{\textbf{H-ELM}} \\ \hline
\textbf{Proposed $\downarrow$} &
  \multicolumn{1}{c|}{p-value} &
  Null hypothesis &
  \multicolumn{1}{c|}{p-value} &
  Null hypothesis &
  \multicolumn{1}{c|}{p-value} &
  Null hypothesis &
  \multicolumn{1}{c|}{p-value} &
  Null hypothesis &
  \multicolumn{1}{c|}{p-value} &
  Null hypothesis \\ \hline
\textbf{F-BLS} &
  \multicolumn{1}{c|}{0.4525} &
  Not Rejected &
  \multicolumn{1}{c|}{0.001234} &
  Rejected &
  \multicolumn{1}{c|}{0.2801} &
  Not Rejected &
  \multicolumn{1}{c|}{{7.94E-06}} &
  Rejected &
  \multicolumn{1}{c|}{0.00951} &
  Rejected \\ \hline
\textbf{IF-BLS} &
  \multicolumn{1}{c|}{0.01917} &
  Rejected &
  \multicolumn{1}{c|}{2.41E-05} &
  Rejected &
  \multicolumn{1}{c|}{0.0009616} &
  Rejected &
  \multicolumn{1}{c|}{{2.81E-06}} &
  Rejected &
  \multicolumn{1}{c|}{1.28E-05} &
  Rejected \\ \hline
\end{tabular}%
}
\end{table*}
\begin{table}[]
\centering
\caption{Pairwise win-tie-loss of proposed and existing models on UCI datasets.}
\label{table:win-tie-loss}
\resizebox{8.9cm}{!}{
\begin{tabular}{|l|cccccc|} \hline
      & \textbf{BLS \cite{chen2017broad}} &
  \textbf{ELM \cite{huang2006extreme}} &
  \textbf{NeuroFBLS \cite{feng2018fuzzy}} &
  \textbf{IF-TSVM \cite{rezvani2019intuitionistic}} &
  \textbf{H-ELM \cite{tang2015extreme}} &
  \textbf{F-BLS $^{\dagger}$} \\ \hline
\textbf{ELM \cite{huang2006extreme}}       & [$3, 3, 22$]  &               &               &              &  &  \\
\textbf{NeuroFBLS \cite{feng2018fuzzy}} & [$10, 2, 16$] & [$15, 4, 9$]  &               &              &  &  \\
\textbf{IF-TSVM \cite{rezvani2019intuitionistic}}   & [$2, 0, 26$]  & [$1, 1, 26$]  & [$0, 2, 26$]  &              &  &  \\
\textbf{H-ELM \cite{tang2015extreme}}     & [$7, 1, 20$]  & [$11, 2, 15$] & [$10, 2, 16$] & [$28, 0, 0$] &  &  \\
\textbf{F-BLS $^{\dagger}$}  & [$9, 5, 14$] & [$21, 3, 4$] & [$16, 2, 10$]      & [$27, 0, 1$]     & [$19, 2, 7$]   &               \\
\textbf{IF-BLS $^{\dagger}$} & [$21, 2, 5$] & [$23, 3, 2$] & [$20, 3, 5$]       & [$28, 0, 0$]     & [$25, 1, 2$]   & [$20, 3, 5$] \\ \hline 
\end{tabular}%
}
\end{table}
\section{Experiments and Results Analysis}
\label{experiments}
To test the efficacy of proposed models, {\em{i.e.,}} F-BLS and IF-BLS, we compare them to baseline models on publicly available UCI \cite{dua2017uci} datasets with and without added Gaussian noise. Moreover, we implement the proposed models on the Alzheimer’s disease (AD) dataset, available on the Alzheimer’s Disease Neuroimaging Initiative (ADNI) ($adni.loni.usc.edu$).
\subsection{Setup for Experiments}
The experimental procedures are executed on a computing system possessing MATLAB R2023a software, an 11th Gen Intel(R) Core(TM) i7-11700 processor operating at 2.50GHz with 16.0 GB RAM, and a Windows-11 operating platform.

Our study involves a comparative analysis among the proposed F-BLS and IF-BLS models and the compared baseline models, namely ELM \cite{huang2006extreme}, BLS \cite{chen2017broad}, NeuroFBLS \cite{feng2018fuzzy}, intuitionistic fuzzy twin support vector machine (IF-TSVM) \cite{rezvani2019intuitionistic} and Hierarchical extreme learned machine (H-ELM) \cite{tang2015extreme}.

We use the 5-fold cross-validation technique and grid search to fine-tune the hyperparameters of models. This involved splitting the dataset into five distinct, non-overlapping folds. For each set of hyperparameters, we calculate the testing accuracy on each fold separately. Average testing accuracy is determined for each set by taking the mean of these five accuracies. The highest average testing accuracy is reported as the accuracy of the models. The hyperparameters setting's range and description are reported in supplementary Table S.I.
\subsection{Results and Statistical Analyses on UCI Dataset}
We select 28 benchmark datasets available in the UCI repository \cite{dua2017uci}, covering diverse domains. We include 14 datasets with sample sizes below 500 and 14 datasets whose sample sizes exceed 500, with feature counts ranging from 4 to 167. For detailed dataset characteristics, please refer to columns 2 and 3 of Table \ref{tab:UCI_Testing_Accuracy}.
 
The performance of these models is assessed using accuracy and is presented in Table \ref{tab:UCI_Testing_Accuracy} and the corresponding standard deviation, rank, and best hyperparameter settings are reported in the supplementary Tables S.III, S.IV, and S.V, respectively. The average accuracies for the existing models are as follows: BLS with an average accuracy of $82.5126\%$, ELM with $79.8093\%$, NeuroFBLS with $80.1096\%$, IF-TSVM with $58.3726\%$, and H-ELM with $79.9287\%$. On the other hand, the proposed IF-BLS and F-BLS models achieved accuracies of $84.5076\%$ and $82.5434\%$, respectively. In terms of accuracy, the proposed IF-BLS and F-BLS ranked first and second, respectively. A notable finding is that the proposed IF-BLS and F-BLS models showed the most minimal standard deviation values among the compared models, with $6.0494$ and $6.1989$, respectively. This shows that the proposed IF-BLS and F-BLS models possess a high degree of certainty in their predictions.

Average accuracy can be a misleading indicator since a model's superior performance on one dataset may make up for a model's inferior performance on another. Therefore, we further utilize a set of statistical metrics, namely, the ranking test, Friedman test, Wilcoxon signed-rank test, and win-tie loss sign test recommended by \citet{demvsar2006statistical}. 
In the ranking scheme, higher ranks are assigned to the worst-performing models and lower to the best-performing models. Suppose $\mathscr{D}$ models are being evaluated using $\mathscr{K}$ datasets, and the $d^{th}$ model's rank on the $k^{th}$ dataset is denoted by $\mathcal{r}(d,k)$. The $d^{th}$ model's average rank is determined as follows:  $\mathcal{r}(d,*)=\left(\sum_{k=1}^{\mathscr{K}}\mathcal{r}(d,k)\right)/\mathscr{K}$. The average ranks of the models are presented in supplementary Table S.IV. The proposed IF-BLS and F-BLS models attained an average rank of $1.8929$ and $3.2679$, respectively. This positions the IF-BLS model as the top-performing model, followed by the F-BLS model, making them the first and third best models, respectively.

The Friedman test compares the average ranks of models and determines whether the models have significant differences based on their rankings. The Friedman test follows the chi-squared distribution ($\chi^2_F$) with $\mathscr{D}-1$ degrees of freedom (d.o.f.) $\chi^2_F = \frac{12\mathscr{K}}{\mathscr{D} (\mathscr{D}+1)} \left(\sum_{d=1}^{\mathscr{D}} \left(\mathcal{r}(d,*)\right)^2 - \frac{\mathscr{D}(\mathscr{D}+1)^2}{4}\right).$
Furthermore, $F_F$ statistic is defined as: $ F_F=\chi_F^2\left(\frac{(\mathscr{K}-1)}{\mathscr{K}(\mathscr{D}-1)-\chi_F^2}\right),$
where the distribution of $F_F$ has $(\mathscr{D}-1)$ and $(\mathscr{K}-1)(\mathscr{D}-1)$ d.o.f..
For $\mathscr{D}=7$ and $\mathscr{K}=28$, we get $\chi^2_F=86.1618$ and $F_F=28.4265$. According to the statistical $F-$distribution table, $F_F (6, 162) = 2.1549$ at $5\%$ level of significance. The null hypothesis is rejected since $28.4265 > 2.1549$. As a result, models differ significantly. Since the null hypothesis is rejected, we use the Wilcoxon signed-rank test \cite{demvsar2006statistical} to assess the pairwise significant distinction among the proposed and baseline models. The Wilcoxon signed-rank test results are presented in Table \ref{tab:UCI_Wilcoxon}; the p-values and the corresponding status of the null hypothesis in pairwise comparisons between the proposed and baseline models underscore a compelling narrative. The consistent rejection of the null hypothesis of the IF-BLS model $w.r.t.$ all the baseline models signifies a statistical superiority of the IF-BLS model, whereas the Wilcoxon signed-rank test confirms that the proposed F-BLS is superior to baseline ELM, IF-TSVM, and H-ELM models. This analysis solidifies the commendable performance and statistical excellence of the proposed F-BLS and IF-BLS models across the board.

In addition, to analyze the models, we use pairwise win-tie-loss sign test. Under the null hypothesis of the win-tie-loss sign test, it is assumed that the models perform equally, $i.e.,$ each model is expected to win on half ($\mathscr{K}/2$) of the datasets out of the total number of datasets ($\mathscr{K}$). For two models to be deemed significantly distinct if one of the models achieves a minimum of $\mathscr{K}/2 + 1.96\sqrt{\mathscr{K}}/2$ victories. In the case of a tie, the score is evenly divided among the models being compared. 
For $\mathscr{K}=28$, the threshold for determining statistical difference according to the win-tie-loss test is equal to $\mathscr{K}/2 + 1.96\sqrt{\mathscr{K}}/2=28/2 + 1.96\sqrt{28}/2=19.1857$. The pairwise win-tie-loss of models are noted in Table \ref{table:win-tie-loss}. If either of the two models achieves victories in a minimum of $20$ datasets, it establishes a clear statistical distinction between them. Upon examination of Table \ref{table:win-tie-loss}, we observe that the proposed IF-BLS model wins over the BLS, ELM, NeuroFBLS, IF-TSVM, H-ELM, and F-BLS models by securing a number of victories: $21$, $23$, $20$, $28$, $25$, and $20$ respectively.  Thereby substantiating the superiority of the proposed IF-BLS model over all the baseline models and proposed F-BLS counterparts. This outcome reinforces the notion that the IF-BLS model possesses notable advantages and stands as a compelling choice in terms of its overall effectiveness.
     
\textbf{\textit{Analysis:}} An intriguing finding emerges from comparing the proposed F-BLS models with the proposed IF-BLS. As expected, the proposed IF-BLS outperforms the proposed F-BLS model, suggesting a plausible explanation. The F-BLS model intelligently handles noise and outliers by assigning fuzzy values to each sample based on membership value considerations. However, relying only on fuzzy membership without incorporating the measure of the extent of non-belongingness to a class is not the optimal way to handle noise and outliers effectively. The remarkable performance of the IF-BLS model offers a compelling solution. By considering both membership and non-membership values when calculating the degree of fuzzy values, the IF-BLS model demonstrates its indispensability in effectively addressing the challenges posed by noise and outliers. This emphasizes how important it is to take into account both membership and non-membership values in order to combat the impact of noise and outliers. 
\begin{table*}[]
\centering
\caption{The classification accuracies of the proposed F-BLS and IF-BLS models along with the existing models, {\em{i.e.,}} BLS, ELM, NeuroFBLS, IF-TSVM, and H-ELM on UCI dataset with varying levels of $5\%$, $10\%$, $15\%$, and $20\%$ Gaussian noise.}
\label{tab:noise_table}
\resizebox{15cm}{!}{%
\begin{tabular}{lcccccccc} \hline \vspace{-3mm}\\ 
\textbf{Dataset $\downarrow$ $\mid$ \text{Model} $\rightarrow$} &
  \textbf{Noise} &
  \textbf{BLS \cite{chen2017broad}} &
  \textbf{ELM \cite{huang2006extreme}} &
  \textbf{NeuroFBLS \cite{feng2018fuzzy}} &
  \textbf{IF-TSVM \cite{rezvani2019intuitionistic}} &
  \textbf{H-ELM \cite{tang2015extreme}} &
  \textbf{F-BLS $^{\dagger}$} &
  \textbf{IF-BLS $^{\dagger}$} \vspace{0mm}\\ \hline \vspace{-3mm}\\
breast\_cancer        & $5\%$  & 70.2359 & 69.4737 & 70.1754 & 70.1754 & \textbf{87.0175} & 71.2704 & 82.807  \\
                      & $10\%$ & 71.9722 & 67.0175 & 70.1754 & 70.1754 & \textbf{83.8838} & 72.6558 & 82.4622 \\
                      & $15\%$ & 70.8832 & 66.6667 & 70.5263 & 70.1754 & 70.5263 & 72.3351 & \textbf{86.6667} \\
                      & $20\%$ & 71.9419 & 67.3926 & 70.5263 & 70.1754 & 76.4912 & 70.5384 & \textbf{81.7604} \vspace{0mm}\\ \hline \vspace{-3mm}\\
conn\_bench\_sonar\_mines\_rocks &
  $5\%$ &
  68.734 &
  57.7003 &
  61.5447 &
  17.9675 &
  73.9141 &
  68.6527 &
  \textbf{78.4204} \\
\textbf{}             & $10\%$ & 67.2009 & 57.2474 & 59.0825 & 18.9547 & 76.8293 & 66.8293 & \textbf{78.2695} \\
                      & $15\%$ & 67.3171 & 67.3403 & 58.2346 & 14.5645 & \textbf{81.2311} & 65.8653 & 77.8513 \\
                      & $20\%$ & 67.7352 & 60.5923 & 61.0221 & 16.0279 & 73.1243 & 68.2695 & \textbf{75.4704} \vspace{0mm}\\ \hline \vspace{-3mm}\\
hill\_valley          & $5\%$  & 81.4359 & 79.0416 & 79.784  & 51.6536 & 74.1764 & \textbf{82.3433} & 81.9311 \\
                      & $10\%$ & \textbf{80.9417} & 76.4847 & 78.1356 & 51.4067 & 71.7832 & 79.4603 & 80.7758 \\
                      & $15\%$ & 80.2809 & 75.579  & 76.7354 & 50.1666 & 70.9693 & 79.3725 & \textbf{80.3639} \\
                      & $20\%$ & 79.2089 & 74.337  & 76.4837 & 50.0027 & 71.2135 & 79.4538 & \textbf{80.0337} \vspace{0mm}\\ \hline \vspace{-3mm}\\
pittsburg\_bridges\_T\_OR\_D &
  $5\%$ &
  89.1905 &
  86.1429 &
  89.2381 &
  86.1429 &
  86.2857 &
  88.1429 &
  \textbf{93.1429} \\
                      & $10\%$ & 88.1905 & 87.2857 & 90.1429 & 86.1429 & 86.1429 & 88.2381 & \textbf{92.1429} \\
                      & $15\%$ & 88.2381 & 86.1905 & 87.2381 & 86.1429 & 86.1429 & 87.1905 & \textbf{92.1429} \\
                      & $20\%$ & 87.2381 & 86.1429 & 92.1429 & 86.1429 & 86.1905 & 90.1429 & \textbf{92.1905} \vspace{0mm}\\ \hline \vspace{-3mm}\\
tic\_tac\_toe         & $5\%$  & \textbf{97.7018} & 81.9181 & 83.3922 & 65.4167 & 67.1673 & 97.284  & 97.3882 \\
                      & $10\%$ & \textbf{97.3893} & 83.5951 & 80.8901 & 65.3125 & 76.3367 & 97.3887 & 97.2846 \\
                      & $15\%$ & 96.0318 & 82.4689 & 78.9998 & 65.3125 & 67.5125 & 95.9271 & \textbf{96.448}  \\
                      & $20\%$ & 94.6755 & 79.2174 & 76.8161 & 65.3125 & 73.6447 & 93.9431 & \textbf{95.8246} \vspace{0mm}\\ \hline \vspace{-3mm}\\
Average Accuracy &
  \multicolumn{1}{l}{} &
  80.8272 &	74.5917	&75.5643	&57.8686	&77.0292	&80.7652	&86.1689 \vspace{0mm}\\ \hline \vspace{-3mm}\\
\multicolumn{9}{l}{The boldface in each row denotes the best-performed model corresponding to each dataset. $\dagger$ represents the proposed models.}
\end{tabular}%
}
\end{table*}
\subsection{Evaluation on UCI Datasets with Gaussian Noise}
While the UCI datasets utilized in our study reflect real-world scenarios, it is worth noting that the presence of impurities or noise in collected data can increase for various reasons. In such circumstances, it becomes imperative to develop a robust model capable of effectively handling such challenging scenarios. To demonstrate the superiority of the proposed F-BLS and IF-BLS models even in adverse situations, we introduce Gaussian noise to selected UCI datasets. We have chosen $5$ diverse UCI datasets for our comparative analysis. 
The reasoning behind selecting the datasets is given in supplementary Section S.II. To conduct a comprehensive analysis, we introduce Gaussian noise with varying levels of $5\%$, $10\%$, $15\%$, and $20\%$ to corrupt the features of these datasets. 

\textbf{\textit{Comparative analysis:}} The accuracy of all the models in the chosen datasets with $5\%$, $10\%$, $15\%$, and $20\%$ noise are presented in Table \ref{tab:noise_table}. Additionally, the standard deviations and best hyperparameters are reported in supplementary Tables S.VI and S.VII, respectively.
\begin{enumerate}
    \item The IF-BLS model consistently achieved the top position in the conn\_bench\_sonar\_mines\_rocks and  pittsburg\_bridges\_T\_OR\_D at $0\%$ noise. Remarkably, it maintained its superior performance even when noise is introduced. In the conn\_bench\_sonar\_mines\_rocks dataset, the IF-BLS model demonstrated the best performance with an accuracy of $75.4704\%$ at $20\%$ noise, which is approx $8\%$ higher than the standard BLS model. We can observe an even better pattern for F-BLS in comparison to BLS. At $0\%$ noise, BLS outperformed the proposed F-BLS, but as the noise increases, the F-BLS starts catching up, and at $20\%$ noise, the F-BLS outperforms the BLS and shows superiority in dealing with noise. Similar results are observed with the  pittsburg\_bridges\_T\_OR\_D dataset.
   \item In Table \ref{tab:noise_table}, we observe interesting patterns in breast\_cancer, hill\_valley, and tic\_tac\_toe datasets. Initially, the proposed F-BLS and IF-BLS models are not the top performers at 0\% noise. However, as the level of noise increased, proposed models began to outperform the baseline models. For instance, in the hill\_valley dataset, the BLS outperforms the proposed models at $0\%$ noise. However, as noise increases, the IF-BLS model becomes the top performer starting from a noise level of $15\%$, while the F-BLS model surpasses BLS at a $20\%$ noise level. Other baseline models do not come close to matching the performance of the proposed models. This trend is consistent across other datasets as well.
   \item The proposed IF-BLS and F-BLS models have secured the first and third positions in terms of average accuracy with values of $86.1689\%$ and $80.7652\%$, respectively. The proposed IF-BLS model's average accuracy is around $6\%$,  $12\%$, $11\%$, $29\%$ and $9\%$ higher than that of baseline models: BLS, ELM, NeuroFBLS, IF-TSVM, H-ELM, respectively. Whereas the proposed F-BLS has around $6\%$,  $5\%$, $23\%$, and $3\%$ higher average accuracy than that of baseline models: ELM, NeuroFBLS, IF-TSVM, H-ELM, respectively. Furthermore, the proposed F-BLS and IF-BLS models exhibit high prediction certainty, with the lowest standard deviations of $9.4511$ and $8.8589$, respectively, compared to the baselines. The combination of the highest average accuracy and the lowest standard deviation indicates that the proposed IF-BLS and F-BLS models are less noise-sensitive and exhibit robust performance in handling varying conditions.
\end{enumerate}
   We subjected the proposed models to challenging conditions and found that the proposed F-BLS and IF-BLS models showcased their robustness and superior performance in unfavorable scenarios. This highlights their adeptness in navigating and excelling in noise-affected environments.
\begin{table*}[]
\centering
\caption{The testing accuracy of the proposed F-BLS and IF-BLS models and the baseline models on ADNI datasets.}
\label{tab:AD_Accuracy}
\resizebox{14cm}{!}{%
\begin{tabular}{lccccccc} \hline \vspace{-3mm}\\ 
        \textbf{Dataset $\downarrow$ $\mid$ \text{Model} $\rightarrow$} &
  \textbf{BLS \cite{chen2017broad}} &
  \textbf{ELM \cite{huang2006extreme}} &
  \textbf{NeuroFBLS \cite{feng2018fuzzy}} &
  \textbf{IF-TSVM \cite{rezvani2019intuitionistic}} &
  \textbf{H-ELM \cite{tang2015extreme}} &
  \textbf{F-BLS $^{\dagger}$} &
  \textbf{IF-BLS $^{\dagger}$} \vspace{0mm}\\ \hline \vspace{-3mm}\\
CN\_VS\_AD  & 88.4337 & 86.747  & 88.6747 & 58.7952 & 82.8916 & 88.6387 & \textbf{88.8137} \\
CN\_VS\_MCI & 70.7695 & 69.0108 & 70.4495 & 63.5733 & 65.6533 & 70.2883 & \textbf{71.0844} \\
MCI\_VS\_AD & 75.0427 & 73.3333 & 74.5299 & 68.0342 & 71.6239 & 74.8718 & \textbf{76.2437} \vspace{0mm}\\ \hline \vspace{-3mm}\\
Average Accuracy  & 78.082  & 76.3637 & 77.8847 & 63.4676 & 73.3896 & 77.9329 & \textbf{78.7139} \vspace{0mm}\\ \hline \vspace{-3mm}\\
\multicolumn{8}{l}{The boldface in each row denotes the best-performed model corresponding to the datasets. $\dagger$ represents the proposed models.}
\end{tabular}%
}
\end{table*}
\begin{figure*}
\begin{minipage}{.32\linewidth}
\centering
\subfloat[CN\_VS\_AD]{\label{main:a}\includegraphics[scale=0.32]{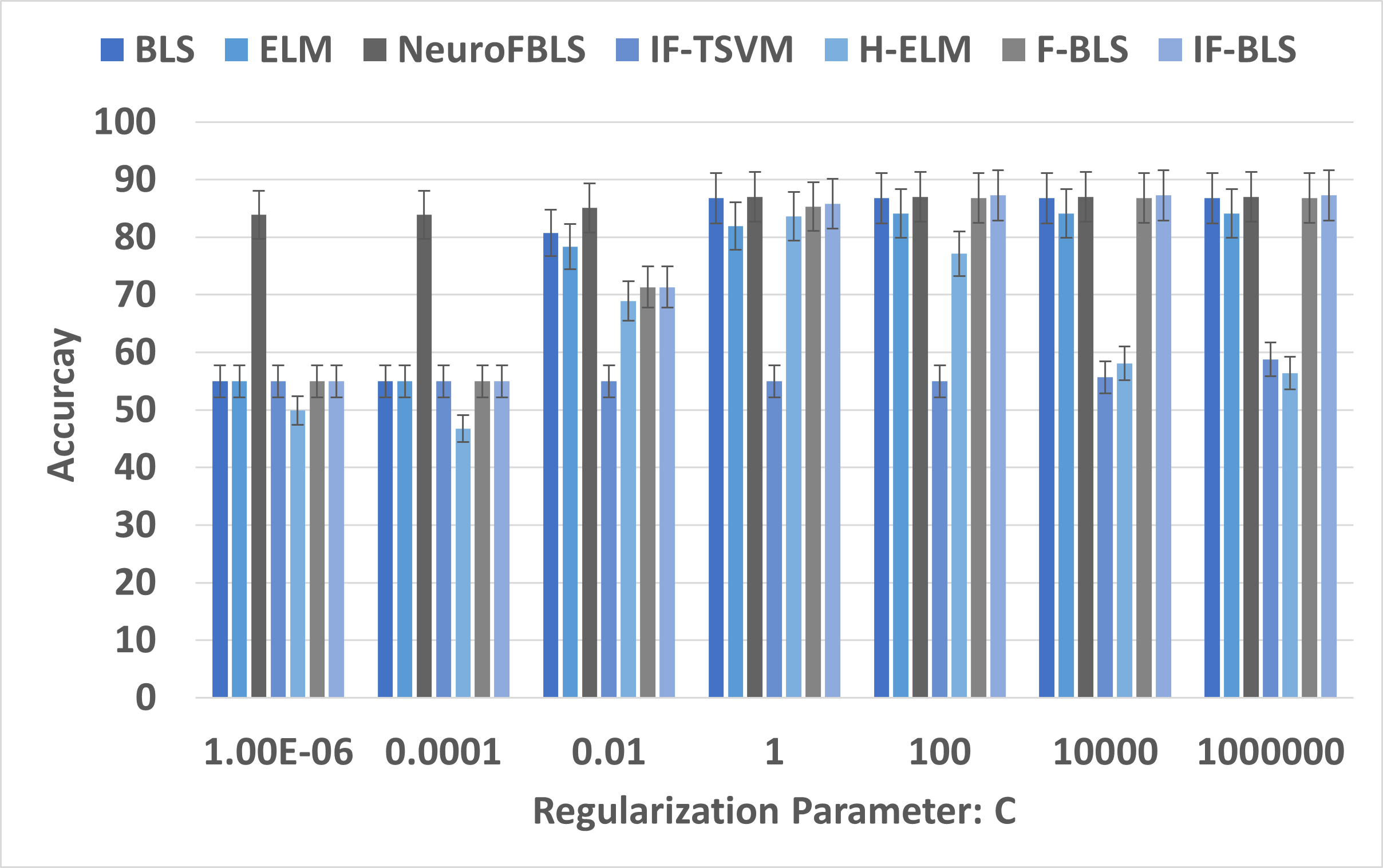}}
\end{minipage}
\begin{minipage}{.34\linewidth}
\centering
\subfloat[CN\_VS\_MCI]{\label{main:b}\includegraphics[scale=0.32]{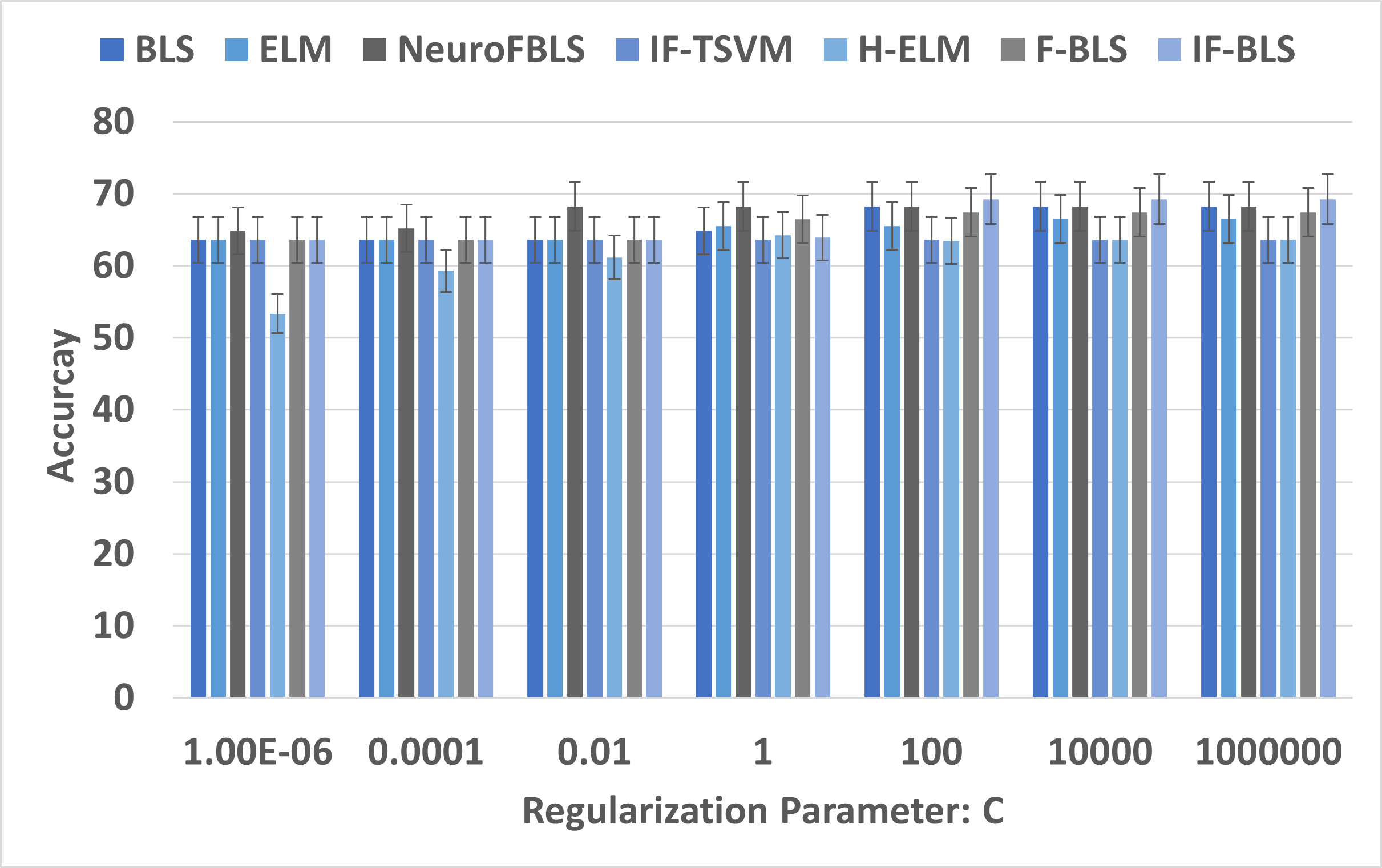}}
\end{minipage}
\begin{minipage}{.33\linewidth}
\centering
\subfloat[MCI\_VS\_AD]{\label{main:c}\includegraphics[scale=0.32]{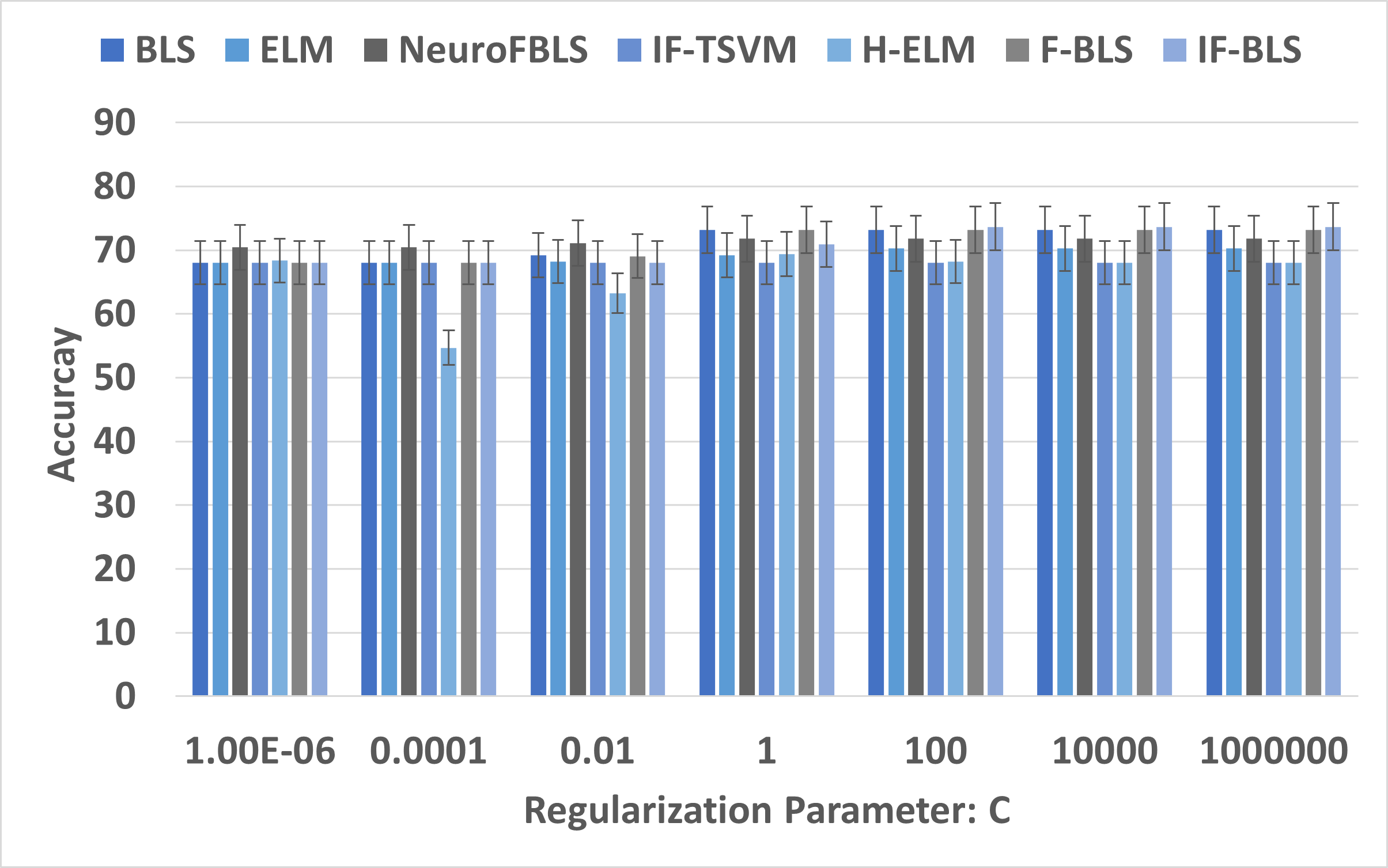}}
\end{minipage}
\par\medskip
\caption{Accuracy along with the standard deviation (shown in thin lines on the top) with respect to regularization parameter $C$ on ADNI dataset.}
\label{fig:AD_Regularization}
\end{figure*}
\subsection{Evaluation on ADNI Dataset}
Alzheimer's Disease (AD) is a degenerative neurological disorder that leads to the gradual deterioration of the brain and is the primary cause of dementia, accounting for $60$-$80\%$ of dementia cases \cite{zhang2020survey}. Predominantly affecting individuals aged $60$ and older, it primarily impairs memory and various cognitive functions. AD can lead to profound changes in behavior, personality, and overall quality of life. Despite decades of research, there is currently no cure for AD, and available treatments only provide temporary symptomatic relief \cite{srivastava2021alzheimer}. Therefore, understanding the complex underlying mechanisms of the disease remains a critical area of investigation to alleviate the burden of AD on individuals, families, and societies.

To train the proposed F-BLS and IF-BLS models, we utilize scans from the Alzheimer's Disease Neuroimaging Initiative (ADNI) dataset, which can be accessed at $adni.loni.usc.edu$. The inception of the ADNI project dates back to $2003$ as a public-private partnership. Wherein its fundamental objective centered around the thorough examination and exploration of diverse neuroimaging techniques, encompassing magnetic resonance imaging (MRI), positron emission tomography (PET), and other diagnostic assessments, to elucidate the intricate nuances of AD at its early stage, characterized by mild cognitive impairment (MCI). The dataset encapsulates three distinctive scenarios, specifically the comparative analysis of control normal (CN) versus MCI (CN\_VS\_MCI), MCI versus AD (MCI\_VS\_AD), and CN versus AD (CN\_VS\_AD ).

Table \ref{tab:AD_Accuracy} shows the accuracies achieved by different models for diagnosing AD, whereas supplementary Table S.VIII reports the best hyperparameter settings for each model.  The proposed IF-BLS excels with a maximum average testing accuracy of $78.7139\%$, whereas the proposed F-BLS secures third position with an average accuracy of $77.9329\%$. The proposed IF-BLS model has the highest accuracy $88.8137\%$ for the CN\_VS\_AD case, followed by NeuroFBLS and F-BLS with accuracy $88.6747\%$ and $88.6387\%$, respectively. For the CN\_VS\_MCI case, the proposed IF-BLS again comes out on top with the average testing accuracy of $71.0844\%$. In the MCI\_VS\_AD scenario, the IF-BLS model achieves the highest accuracy of $76.2437\%$, while the F-BLS model follows closely with an accuracy of $74.8718\%$, securing third place. In the overall comparison, the proposed IF-BLS model emerges as a top-performing model by consistently achieving high accuracies across different cases, and the proposed F-BLS shows a competitive nature. These findings highlight the effectiveness of the IF-BLS and F-BLS models in accurately distinguishing individuals with CN, MCI, and AD cases. 
\subsection{Sensitivity Analyses of the Hyperparameters of the Proposed Models}
To understand the behavior of the proposed models, it is essential to understand the dependence of models on their hyperparameters. Thus, we conduct the following sensitivity analyses of the hyperparameters to gain a comprehensive perspective of the proposed models: (i) investigating the impact of varying the number of regularization parameters ($C$) on model performance, (ii) assessing the dependence of the proposed IF-BLS models on Feature Groups ($m$), kernel parameter ($\mu$), and $C$, (iii) exploring the influence of the number of feature nodes ($p$) on both F-BLS and IF-BLS models.

To Investigate the impact of varying the number of regularization parameters ($C$), we draw Figure \ref{fig:AD_Regularization}, which shows the testing accuracy along with the standard deviation (shown in thin lines on the top) of each model with respect to the regularization parameter $C$ on the ADNI dataset. For each case, we fix the other hyperparameters equal to the best values, which are listed in supplementary Table S.VIII. From Figure \ref{fig:AD_Regularization}, we observe that, with the increase in values of $C$, testing accuracies of each model are also increasing. At $C=10^6$, IF-BLS outperforms other models with the highest testing accuracy in each case. The rest of the sensitivity analyses are discussed in Section S.III of the supplementary material.
 \section{Conclusion and Future Work}
 \label{conclusion}
 The BLS model is negatively impacted by noise and outliers present in the datasets, which leads to reduced robustness and potentially biased outcomes during the training phase. To address this issue, we propose the fuzzy BLS (F-BLS) and intuitionistic fuzzy BLS (IF-BLS) models. The proposed F-BLS model considers the distance from samples to the class center in the sample space while assigning weight to each sample. Meanwhile, IF-BLS assigns scores to training points in the kernel space. The effectiveness of the proposed F-BLS and IF-BLS models is demonstrated by applying them to standard benchmark datasets obtained from the UCI repository. Through statistical analyses such as average accuracy, ranking scheme, Friedman test, Wicoxon-signed rank test, and win-tie-loss sign test, both the proposed F-BLS and IF-BLS models exhibit superiority over baseline models. To test the robustness of the proposed F-BLS and IF-BLS models, we introduced Gaussian noise to various UCI datasets. Results indicate that the proposed models exhibit accuracy improvements ranging from $3\%$ to $29\%$ compared to baseline models. The proposed F-BLS and IF-BLS models are also employed for AD diagnosis. The proposed F-BLS shows a competitive nature and the IF-BLS model achieved the highest average accuracy as well as the highest testing accuracy across all three cases (AD\_VS\_MCI, AD\_VS\_CN, and CN\_VS\_MCI). In line with the research findings \cite{tanveer2020machine}, the MCI\_VS\_AD case is identified as the most challenging case in AD diagnosis. However, the proposed F-BLS and IF-BLS models attained the top positions with testing accuracy of $74.8718\%$ and $76.2437\%$, respectively. The proposed F-BLS and IF-BLS models consistently show lower standard deviations compared to baseline models, indicating their higher certainty and robust performance across diverse conditions.
However, because of the kernel function in the IFM scheme, the parameter $\mu$ gets added to the set of the model's tunable parameter and increases the computational burden on the proposed IF-BLS model. Exploring the extension of the proposed models by proposing an IFM scheme without any additional parameters is an interesting avenue for future work. Source code link of the proposed model: \emph{https://github.com/mtanveer1/IF-BLS}.
 \section*{Acknowledgment}
 \label{acknowledge}
This project is supported by the Indian government's Department of Science and Technology (DST) and Ministry of Electronics and Information Technology (MeitY) through grant no. DST/NSM/R\&D\_HPC\_Appl/2021/03.29 under National Supercomputing Mission scheme and Science and Engineering Research Board (SERB) grant no. MTR/2021/000787 under Mathematical Research Impact-Centric Support (MATRICS) scheme.
The Council of Scientific and Industrial Research (CSIR), New Delhi, provided a fellowship for Md Sajid's research under the grant no. 09/1022(13847)/2022-EMR-I.
The Alzheimer's Disease Neuroimaging Initiative (ADNI), which was funded by the Department of Defense's ADNI contract W81XWH-12-2-0012 and the National Institutes of Health's U01 AG024904 grant, allowed for the acquisition of the dataset used in this work. The National Institute on Ageing, the National Institute of Biomedical Imaging and Bioengineering, and other generous donations from a range of organizations provided money for the aforementioned initiative: F. Hoffmann-La Roche Ltd. and its affiliated company Genentech, Inc.; Bristol-Myers Squibb Company;  Alzheimer’s Drug Discovery Foundation; Merck \& Co., Inc.; Johnson \& Johnson Pharmaceutical Research \& Development LLC.;  NeuroRx Research;   Novartis Pharmaceuticals Corporation; AbbVie, Alzheimer’s Association; CereSpir, Inc.;   IXICO Ltd.; Araclon Biotech; BioClinica, Inc.; Lumosity; Biogen; Fujirebio; EuroImmun; Piramal Imaging; GE Healthcare; Cogstate; Meso Scale Diagnostics, LLC.;   Servier; Eli Lilly and Company; Transition Therapeutics Elan Pharmaceuticals, Inc.; Janssen Alzheimer Immunotherapy Research \& Development, LLC.; Lundbeck; Eisai Inc.; Neurotrack Technologies; Pfizer Inc. and Takeda Pharmaceutical Company. The maintenance of ADNI clinical sites across Canada is being funded by the Canadian Institutes of Health Research. In the meanwhile, private sector donations have been made possible to fund this effort through the Foundation for the National Institutes of Health (www.fnih.org).  The Northern California Institute and the University of Southern California's Alzheimer's Therapeutic Research Institute provided funding for the awards intended for research and teaching. The Neuro Imaging Laboratory at the University of Southern California was responsible for making the ADNI initiative's data public. The ADNI dataset available at adni.loni.usc.edu is used in this investigation. 
\bibliographystyle{IEEEtranN}
 \bibliography{refs}
\end{document}


\title{Supplementary Material for the Manuscript ``Intuitionistic Fuzzy Broad Learning System: Enhancing Robustness Against Noise and Outliers"\\}

\author{M. Sajid, A.K. Malik, M. Tanveer, for the Alzheimer’s Disease Neuroimaging Initiative} 

\markboth{  }%
{Shell \MakeLowercase{\textit{et al.}}: A Sample Article Using IEEEtran.cls for IEEE Journals}


\maketitle
\section{Kernel technique used for finding radius in assigning intuitionistic fuzzy score}
The Kernel technique is explored here.
\begin{strip}
\begin{theorem}
\cite{ha2013support}: Let $\mathcal{K}(x_r, x_l)= \psi(x_r).\psi(x_l)$ be a Kernel function, where $.$ is the standard inner product. Then, the norm $\|\cdot\|$ is calculated as:
\begin{equation}
    \|\psi(x_r)-\psi(x_l)\|=\sqrt{\mathcal{K}(x_r, x_r) + \mathcal{K}(x_l, x_l) - 2\mathcal{K}(x_r, x_l)}.
\end{equation}
\end{theorem}
\begin{proof}
\begin{align*}
 \|\psi(x_r)-\psi(x_l)\|^2&=\big(\psi(x_r)-\psi(x_l)\big).\big(\psi(x_r)-\psi(x_l)\big)\\
 &=\psi(x_r).\psi(x_r)+\psi(x_l).\psi(x_l)-2\psi(x_r).\psi(x_l)\\
 &=\mathcal{K}(x_r, x_r)+\mathcal{K}(x_l, x_l)- 2\mathcal{K}(x_r, x_l).
 \end{align*}
\end{proof}
\begin{theorem}
\cite{ha2013support}: The radii $R_{pos}$ and $R_{neg}$ are calculated as:
\begin{equation}
    R_{pos}=\max_{t_r=+1}\sqrt{\mathcal{K}(x_r, x_r) + \frac{1}{N_{pos}^2} \sum_{t_i=+1}\sum_{t_j=+1}\mathcal{K}(x_i, x_j) - \frac{2}{N_{pos}}\sum_{t_l=+1}\mathcal{K}(x_r, x_l)}
\end{equation}
\begin{equation}
    R_{neg}=\max_{t_r=-1}\sqrt{\mathcal{K}(x_r, x_r) + \frac{1}{N_{neg}^2} \sum_{t_i=-1}\sum_{t_j=-1}\mathcal{K}(x_i, x_j) - \frac{2}{N_{neg}}\sum_{t_l=-1}\mathcal{K}(x_r, x_l)}.
\end{equation}
\end{theorem}
\begin{proof}
\begin{align*}
R_{pos}&= \max_{t_r=+1}\|\psi(x_r)-C_{pos}\|\\
&=\max_{t_r=+1}\sqrt{\big(\psi(x_r)-C_{pos}\big).\big(\psi(x_r)-C_{pos}\big)}\\
&=\max_{t_r=+1}\sqrt{\psi(x_r).\psi(x_r)+C_{pos}.C_{pos}-2\psi(x_r).C_{pos}}\\
 &=\max_{t_r=+1}\sqrt{\mathcal{K}(x_r, x_l)+\Bigg(\frac{1}{N_{pos}}\sum_{t_i=+1}\psi(x_i)\Bigg)\Bigg(\frac{1}{N_{pos}}\sum_{t_j=+1}\psi(x_j)\Bigg)-2(\psi(x_i))\Bigg(\frac{1}{N_{pos}}\sum_{t_j=+1}\psi(x_j)\Bigg)}\\
 &=\max_{t_r=+1}\sqrt{\mathcal{K}(x_r, x_r) + \frac{1}{N_{pos}^2} \sum_{t_i=+1}\sum_{t_j=+1}\mathcal{K}(x_i, x_j) - \frac{2}{N_{pos}}\sum_{t_l=+1}\mathcal{K}(x_r, x_l)}.
 \end{align*}
 Similarly, the radius $R_{neg}$ can be calculated.
\end{proof}
\end{strip}
\begin{table*}[htp]
\centering
\caption{Hyperparameters' description and range for the baseline and proposed models.}
\label{tab:Randomized_Parameters_Range}
\resizebox{1.0\linewidth}{!}{
\begin{tabular}{lll}
\hline
\multicolumn{1}{|l|}{Models} & \multicolumn{1}{c|}{Parameters' description} & \multicolumn{1}{c|}{Parameters' Range} \\ \hline
\multicolumn{1}{|l|}{BLS}    & \multicolumn{1}{l|}{\begin{tabular}[c]{@{}l@{}}$C$: Regularization parameter\\$ m$: Number of feature groups\\$p$: Number of feature nodes in each groups\\$q$: Number of enhancement nodes\end{tabular}}       & \multicolumn{1}{l|}{\begin{tabular}[c]{@{}l@{}}$C=[10^{-6}, 10^{-4},\cdots,10^6]$\\$ m=1:2:21$\\$p=5:5:50$\\$q=5:10:105$\end{tabular}}\\ \hline
\multicolumn{1}{|l|}{ELM} &
  \multicolumn{1}{l|}{\begin{tabular}[c]{@{}l@{}}$C$: Regularization parameter\\ $h_l$: Number of hidden nodes\end{tabular}} &
  \multicolumn{1}{l|}{\begin{tabular}[c]{@{}l@{}} $C=[10^{-6}, 10^{-4},\cdots,10^6]$\\ $h_l=5:10:205$\end{tabular}} \\ \hline
\multicolumn{1}{|l|}{NeuroFBLS}    & \multicolumn{1}{l|}{\begin{tabular}[c]{@{}l@{}}$C$: Regularization parameter\\$ N_{fg}$: Number of fuzzy groups\\$N_{fn}$: Number of fuzzy nodes in each groups\\$q$: Number of enhancement nodes\end{tabular}}       & \multicolumn{1}{l|}{\begin{tabular}[c]{@{}l@{}}$C=[10^{-6}, 10^{-4},\cdots,10^6]$\\$ N_{fg}=1:2:21$\\$N_{fn}=5:5:50$\\$q=5:10:105$\end{tabular}}\\ \hline
\multicolumn{1}{|l|}{IF-TSVM} &
  \multicolumn{1}{l|}{\begin{tabular}[c]{@{}l@{}}$C_1$: Regularization parameter for the positive class\\ $C_2$: Regularization parameter for the negative class \\ $\mu$: Intuitionistic fuzzy Kernel parameter \end{tabular}} &
  \multicolumn{1}{l|}{\begin{tabular}[c]{@{}l@{}}$C_1=[10^{-6}, 10^{-4},\cdots,10^6]$  \\ $C_2=[10^{-6}, 10^{-4},\cdots,10^6]$ \\ $\mu=[2^{-5},2^{-4},\cdots,2^{5}]$\end{tabular}} \\ \hline
\multicolumn{1}{|l|}{H-ELM} &
  \multicolumn{1}{l|}{\begin{tabular}[c]{@{}l@{}}$C$: Regularization parameter\\ $h_l$: Number of hidden nodes\end{tabular}} &
  \multicolumn{1}{l|}{\begin{tabular}[c]{@{}l@{}} $C=[10^{-6}, 10^{-4},\cdots,10^6]$\\ $h_l=5:10:205$\end{tabular}} \\ \hline
\multicolumn{1}{|l|}{F-BLS}    & \multicolumn{1}{l|}{\begin{tabular}[c]{@{}l@{}}$C$: Regularization parameter\\$ m$: Number of feature groups\\$p$: Number of feature nodes in each groups\\$q$: Number of enhancement nodes\end{tabular}}       & \multicolumn{1}{l|}{\begin{tabular}[c]{@{}l@{}}$C=[10^{-6}, 10^{-4},\cdots,10^6]$\\$ m=1:2:21$\\$p=5:5:50$\\$q=5:10:105$\end{tabular}}\\ \hline
\multicolumn{1}{|l|}{IF-BLS}    & \multicolumn{1}{l|}{\begin{tabular}[c]{@{}l@{}}$C$: Regularization parameter\\$ m$: Number of fuzzy groups\\$p$: Number of fuzzy nodes in each groups\\$q$: Number of enhancement nodes \\ $\mu$: Intuitionistic fuzzy Kernel parameter\end{tabular}}       & \multicolumn{1}{l|}{\begin{tabular}[c]{@{}l@{}}$C=[10^{-6}, 10^{-4},\cdots,10^6]$\\$ m=1:2:21$\\$p=5:5:50$\\$q=5:10:105$\\ $\mu=[2^{-5},2^{-4},\cdots,2^{5}]$\end{tabular}}\\ \hline
\end{tabular}
}
\end{table*}
\begin{table*}[htp]
\centering
\caption{Description of UCI datasets chosen for experiments with added Gaussian noise.}
\resizebox{11cm}{!}{
\label{tab:description_datasets_noise}
\begin{tabular}{|l|c|c|}
\hline
Dataset                      & Subject Area      & Win/Loss of the proposed   \\  &           &   IF-BLS model in 0\% noise                      \\\hline
breast\_cancer                & Life              & Loss                           \\
conn\_bench\_sonar\_mines\_rocks & Physical          & Win                         \\
hill\_valley                  & Other (Synthetic) & Loss                        \\
pittsburg\_bridges\_T\_OR\_D     & Civil Engineering & Win                           \\
tic\_tac\_toe                  & Game              & Loss                        \\ \hline
\end{tabular}}
\end{table*}
\section{Datasets selected for evaluation on UCI datasets with Gaussian noise}
We have chosen $5$ diverse UCI datasets for our comparative analysis to evaluate the proposed F-BLS and IF-BLS models in the presence of Gaussian noise. These datasets include breast\_cancer, conn\_bench\_sonar\_mines\_rocks, hill\_valley, pittsburg\_bridges\_T\_OR\_D, and tic\_tac\_toe, each representing a different domain. In order to ensure fairness in evaluating the models, we have selected $3$ datasets where none of the proposed F-BLS and IF-BLS models achieve the highest performance at the noise level of $0\%$ (indicating a loss for the proposed models) as per Table I of the main file of the paper. Additionally, we have chosen $2$ datasets where the proposed F-BLS or IF-BLS model outperforms other models at the $0\%$ noise level (indicating a win for the proposed models). For detailed information regarding the selected datasets, please refer to Table S.\ref{tab:description_datasets_noise}.
\section{Sensitivity Analysis of the Proposed F-BLS and IF-BLS models}
In this section, we further analyze the dependency of the proposed models on the hyperparameters.
\subsection{Performance Dependence on Feature Groups ($m$), the Kernel Parameter ($\mu$) and Regularization Parameter ($C$)} 
\begin{figure*}
\begin{minipage}{.5\linewidth}
\centering
\subfloat[$m=3$]{\label{main:a}\includegraphics[scale=0.64]{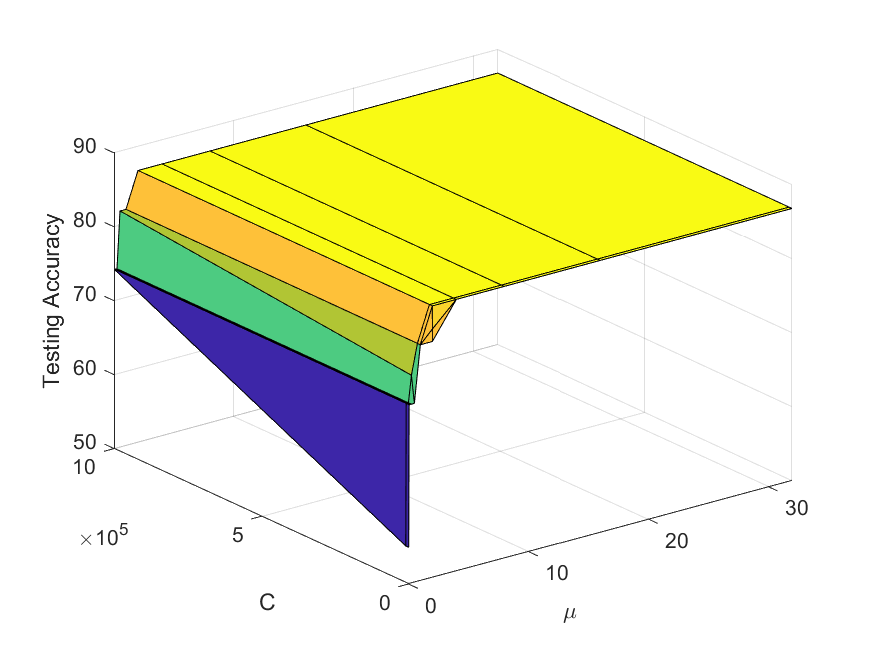}}
\centering
\subfloat[$m=9$]{\label{main:b}\includegraphics[scale=0.64]{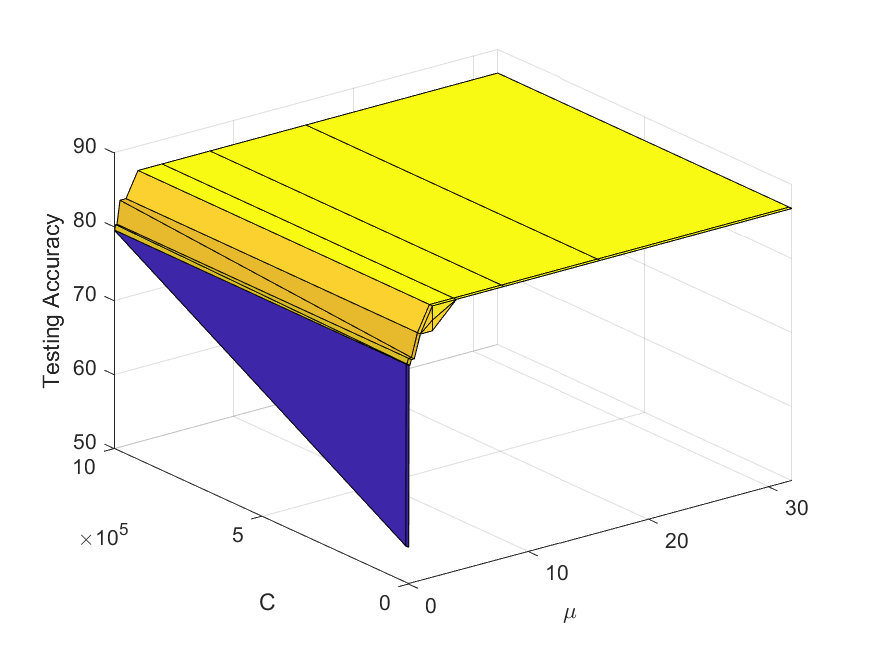}}
\end{minipage}\par\medskip
\begin{minipage}{.5\linewidth}
\centering
\subfloat[$m=15$]{\label{main:c}\includegraphics[scale=0.64]{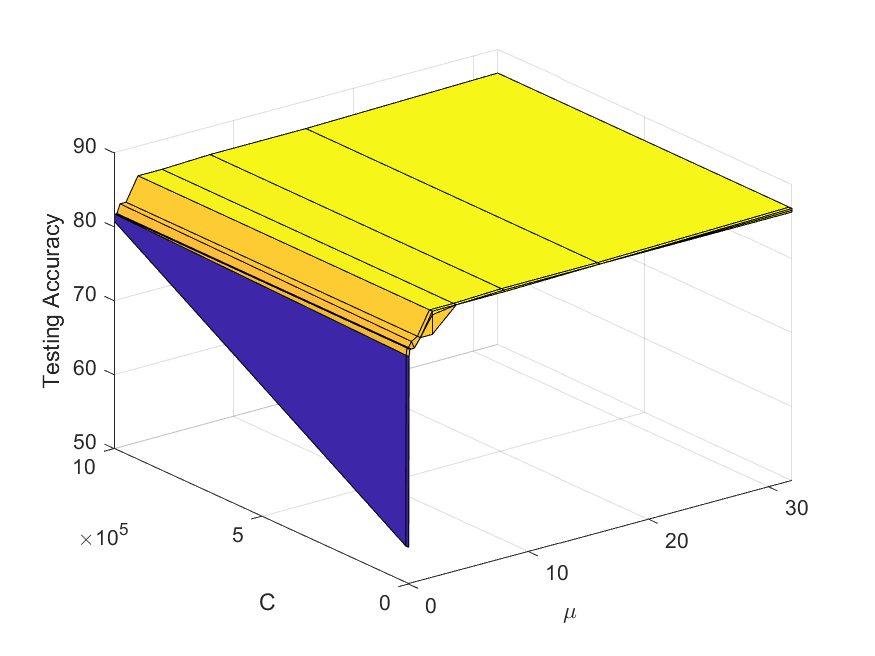}}
\centering
\subfloat[$m=21$]{\label{main:b}\includegraphics[scale=0.64]{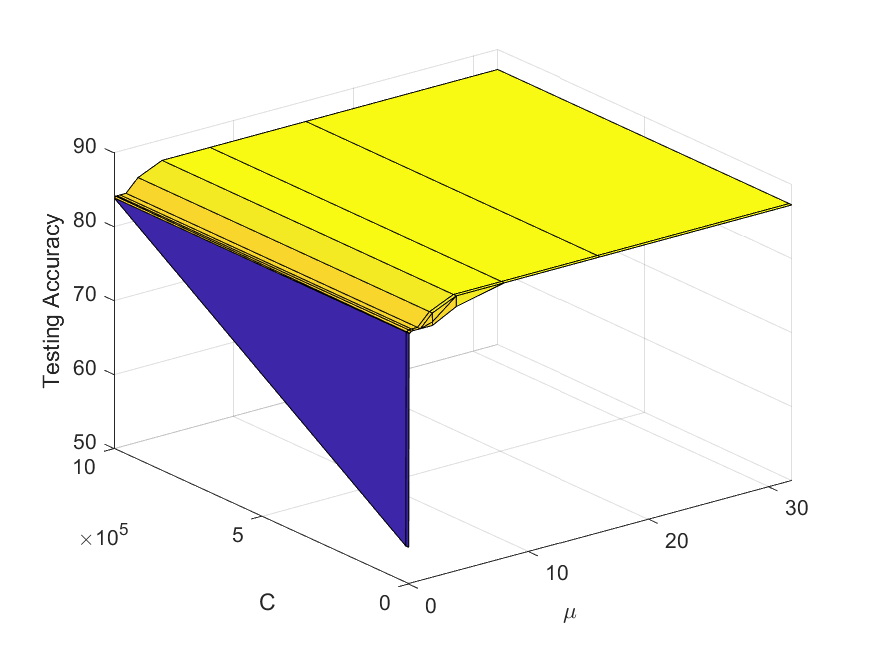}}
\end{minipage}\par\medskip
\caption{Evaluation of the proposed IF-BLS model's testing accuracy on the CN\_VS\_AD case for Alzheimer's disease diagnosis, considering different numbers of feature groups ($m$), while varying the Kernel parameter ($\mu$) and regularization parameter ($C$).}
\label{fig:CN_VS_AD}
\end{figure*}

\begin{figure*}
\begin{minipage}{.5\linewidth}
\centering
\subfloat[$m=3$]{\label{main:a}\includegraphics[scale=0.64]{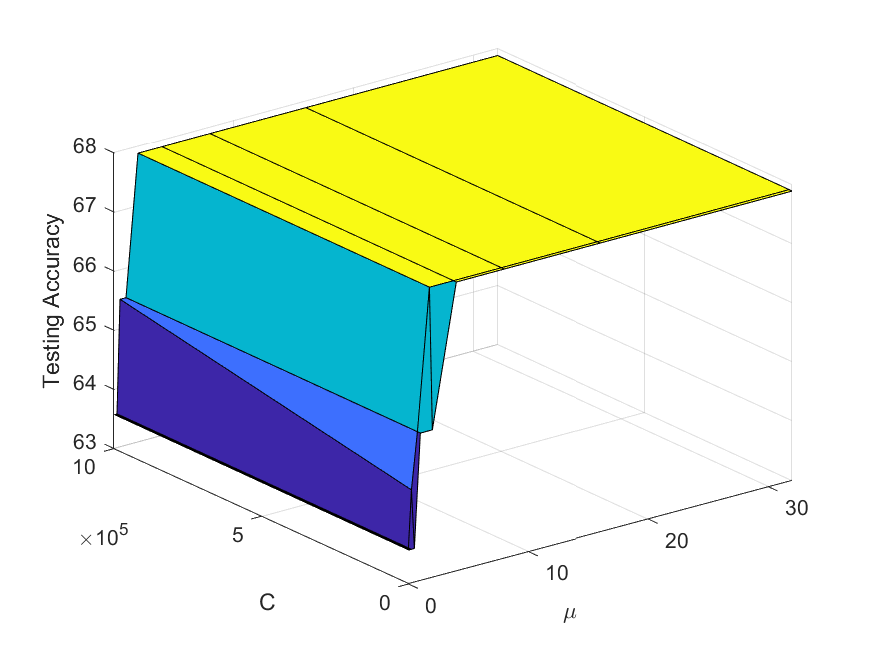}}
\centering
\subfloat[$m=9$]{\label{main:b}\includegraphics[scale=0.64]{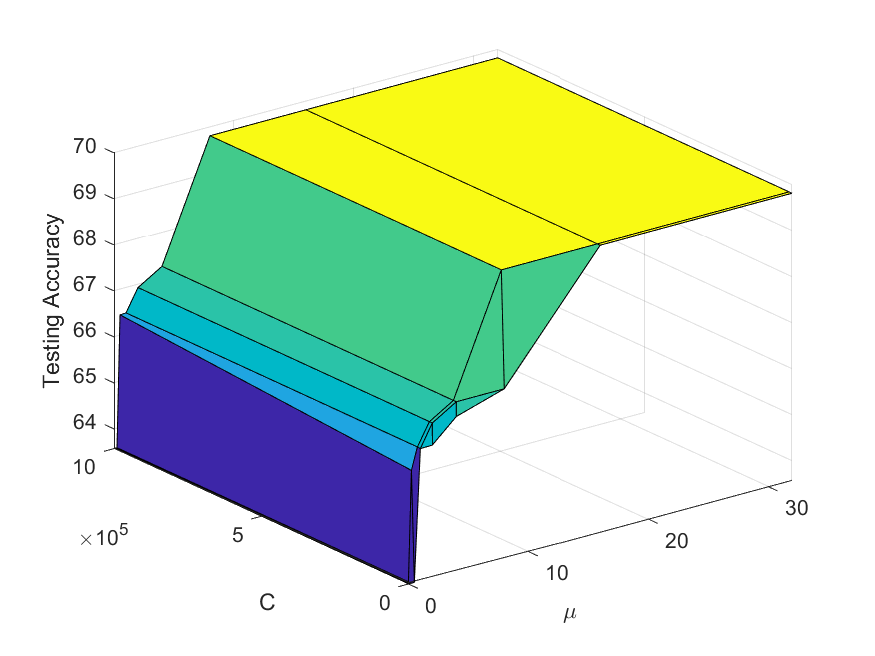}}
\end{minipage}\par\medskip
\begin{minipage}{.5\linewidth}
\centering
\subfloat[$m=15$]{\label{main:c}\includegraphics[scale=0.64]{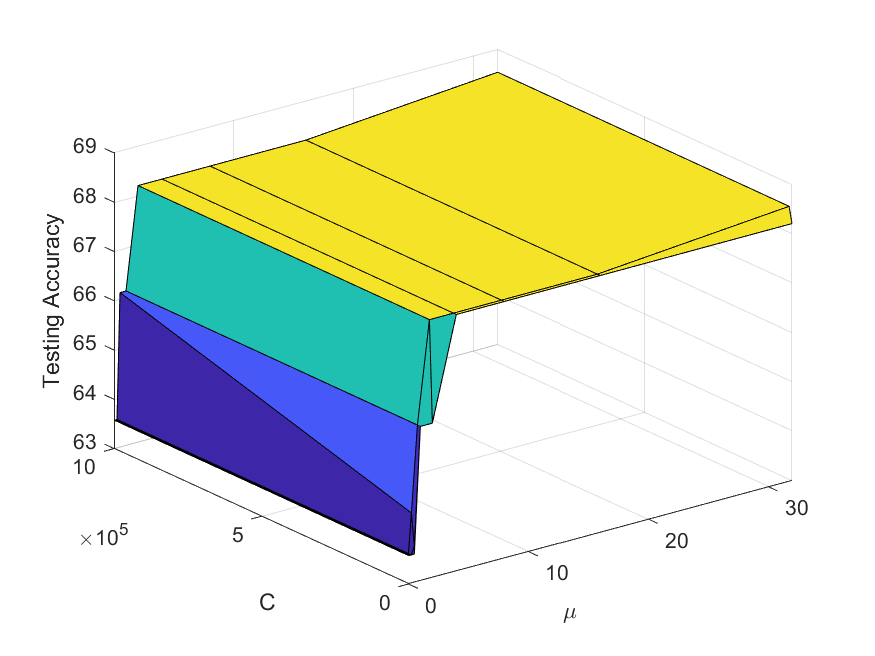}}
\centering
\subfloat[$m=21$]{\label{main:b}\includegraphics[scale=0.64]{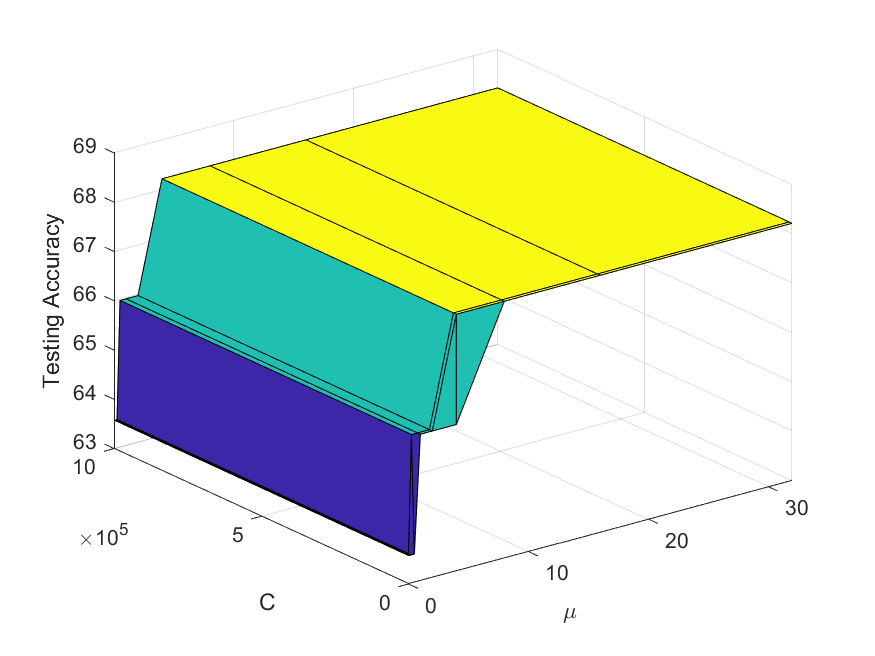}}
\end{minipage}\par\medskip
\caption{Evaluation of the proposed IF-BLS model's testing accuracy on the CN\_VS\_MCI case for Alzheimer's disease diagnosis, considering different numbers of feature groups ($m$), while varying the Kernel parameter ($\mu$) and regularization parameter ($C$).}
\label{fig:CN_VS_MCI}
\end{figure*}

\begin{figure*}
\begin{minipage}{.5\linewidth}
\centering
\subfloat[$m=3$]{\label{main:a}\includegraphics[scale=0.64]{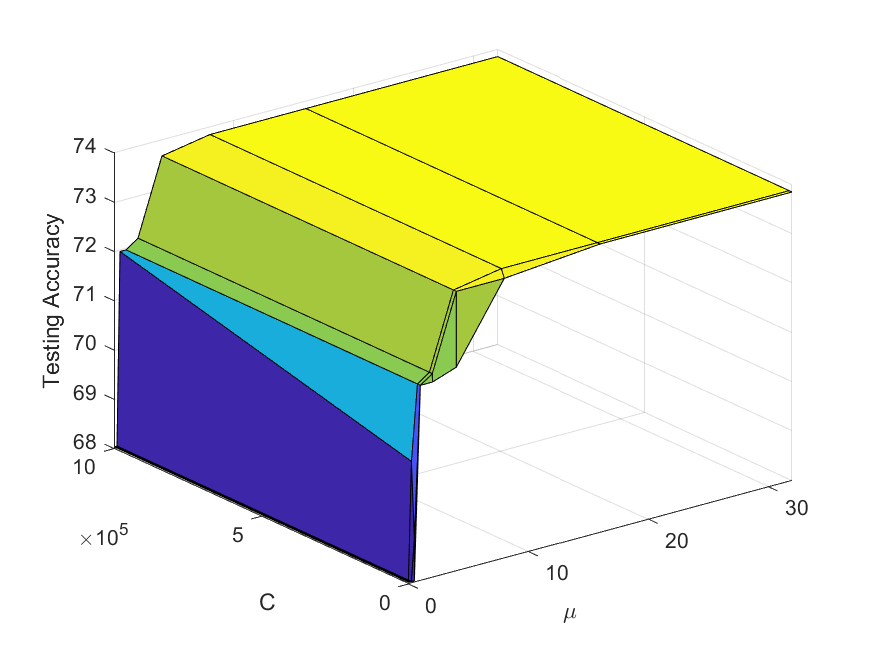}}
\centering
\subfloat[$m=9$]{\label{main:b}\includegraphics[scale=0.64]{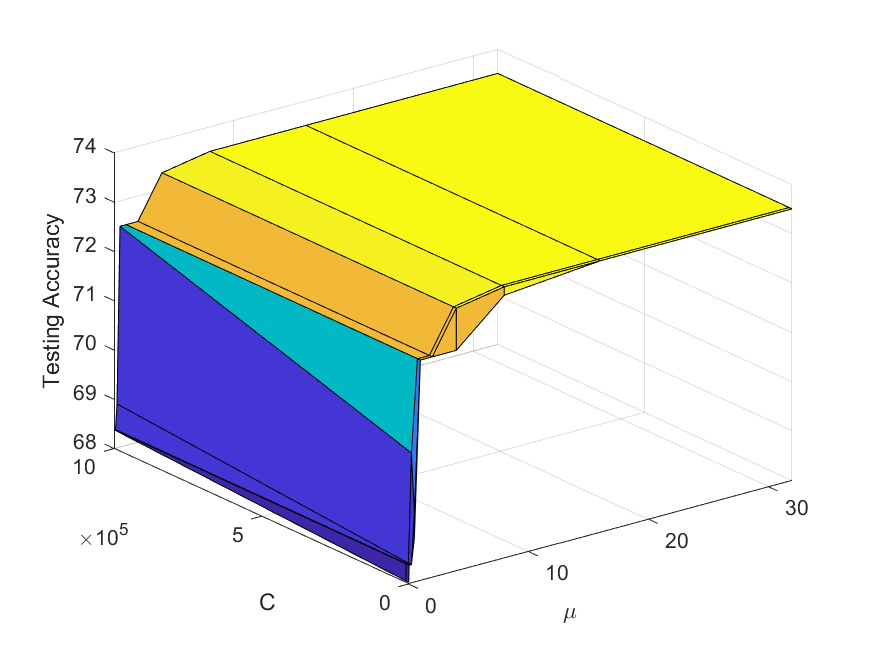}}
\end{minipage}\par\medskip
\begin{minipage}{.5\linewidth}
\centering
\subfloat[$m=15$]{\label{main:c}\includegraphics[scale=0.64]{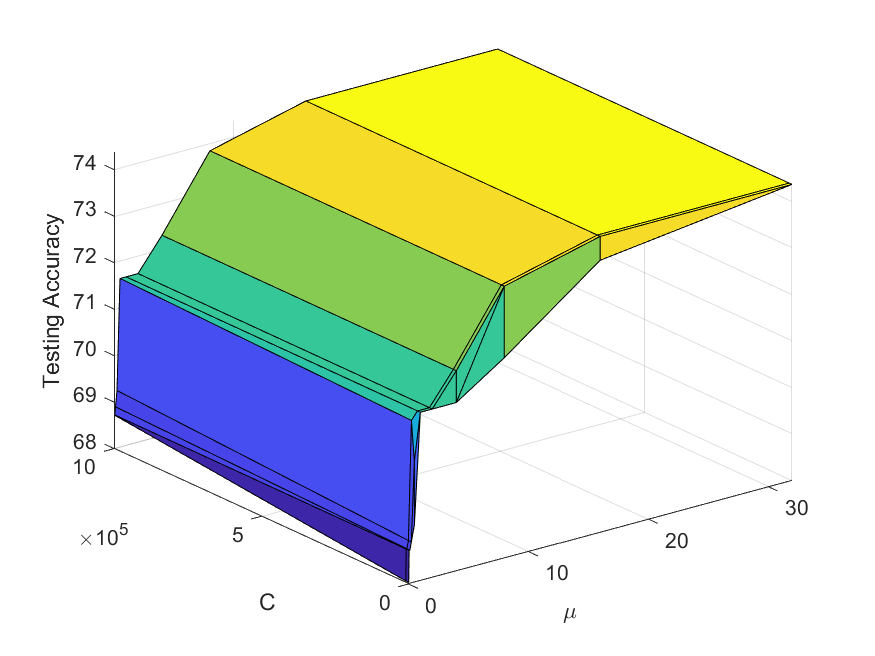}}
\centering
\subfloat[$m=21$]{\label{main:b}\includegraphics[scale=0.64]{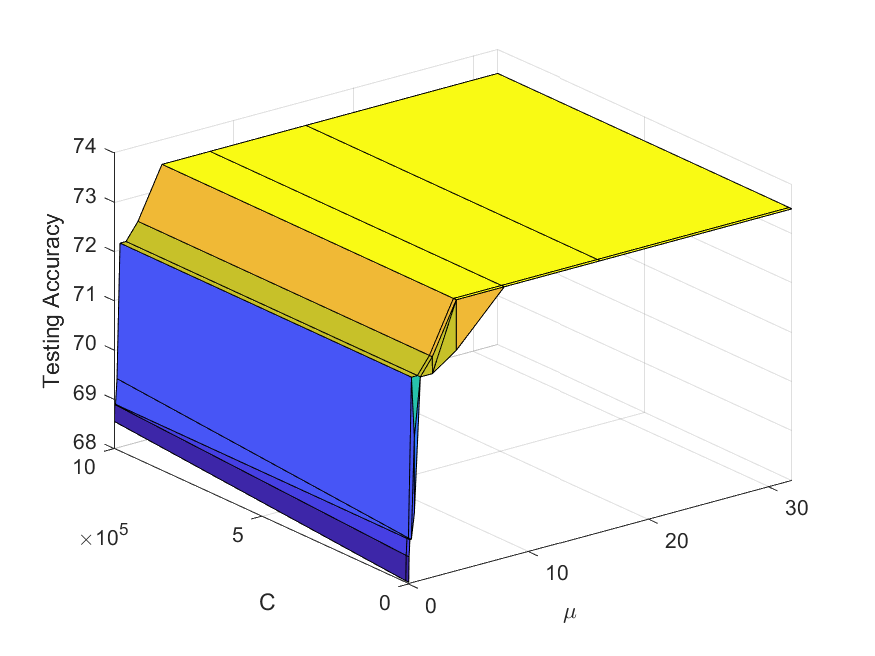}}
\end{minipage}\par\medskip
\caption{Evaluation of the proposed IF-BLS model's testing accuracy on the MCI\_VS\_AD case for Alzheimer's disease diagnosis, considering different numbers of feature groups ($m$), while varying the Kernel parameter ($\mu$) and regularization parameter ($C$).}
\label{fig:MCI_VS_AD}
\end{figure*}

The testing accuracy of the proposed IF-BLS model for Alzheimer's disease diagnosis is depicted in Figures S.\ref{fig:CN_VS_AD}, S.\ref{fig:CN_VS_MCI}, and S.\ref{fig:MCI_VS_AD} for the CN\_VS\_AD, CN\_VS\_MCI, and MCI\_VS\_AD cases, respectively. The accuracy is evaluated by considering different numbers of feature groups ($m$) while varying the Kernel parameter ($\mu$) and regularization parameter ($C$). One can observe that as the values of $\mu$ and $C$ increase, the accuracy also increases. However, beyond a certain point, the accuracy plateau is reached, indicating that the accuracy becomes less responsive to increases in $\mu$ and $C$. Therefore, more attention is required to choose the hyperparameters.
\subsection{Performance Dependence on Number of Feature Nodes ($p$)} 
\begin{figure*}
\begin{minipage}{.5\linewidth}
\centering
\subfloat[F-BLS]{\label{main:a}\includegraphics[scale=0.37]{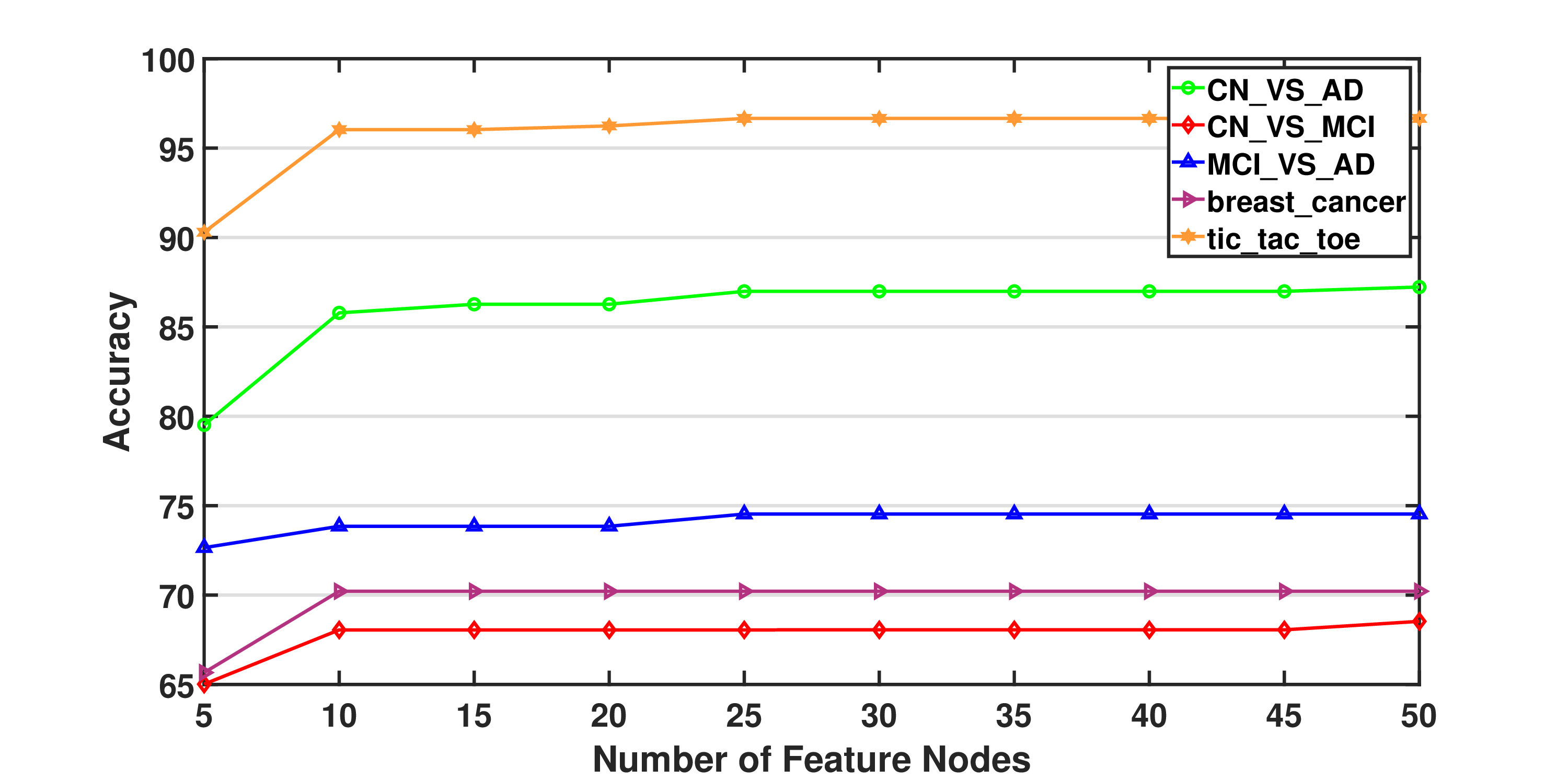}}
\end{minipage}\par\medskip
\begin{minipage}{.5\linewidth}
\centering
\subfloat[IF-BLS]{\label{main:b}\includegraphics[scale=0.37]{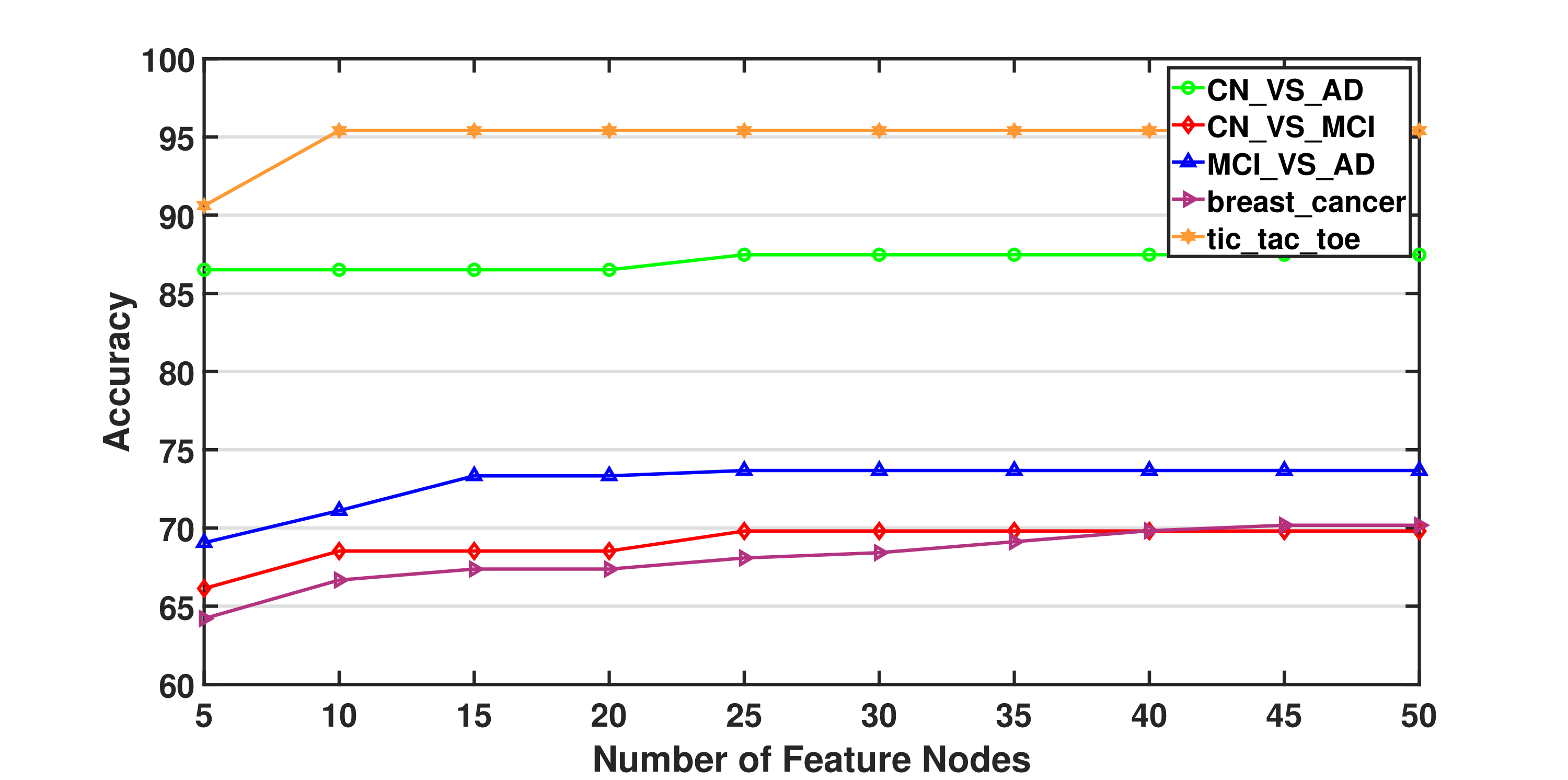}}
\end{minipage}\par\medskip
\caption{Evaluation of the proposed F-BLS and IF-BLS models' testing accuracy while varying the number of feature nodes ($p$).}
\label{fig:Feature_Nodes}
\end{figure*}

Moreover, to assess the impact of varying the number of feature nodes on the performance of the proposed F-BLS and IF-BLS models, Figure S.\ref{fig:Feature_Nodes} illustrates the relationship between the number of feature nodes and accuracy across five datasets: CN\_VS\_AD, CN\_VS\_MCI, MCI\_VS\_AD, breast\_cancer, and tic\_tac\_toe. The figure shows that increasing the number of feature nodes initially boosts accuracy. However, once a certain point is reached, increasing the number of feature nodes doesn't lead to much improvement in accuracy.
    
Specifically, for the F-BLS model, optimal accuracy is attained with just $25$ nodes across all datasets. Conversely, the IF-BLS model exhibits varying patterns: for CN\_VS\_AD, tic\_tac\_toe, and MCI\_VS\_AD datasets, peak accuracy is also achieved with $25$ nodes, whereas for CN\_VS\_MCI and breast\_cancer datasets, the optimal accuracy is achieved with $40$ nodes. Consequently, a general recommendation emerges: fine-tuning the number of feature nodes from $25$ onwards offers a promising avenue towards achieving optimal accuracy. However, it's important to note that the model's performance is dataset-dependent, necessitating careful hyperparameter tuning for each dataset to maximize accuracy.
\begin{table*}[htp]
\centering
\caption{The standard deviations of the proposed F-BLS and IF-BLS models along with the existing models, {\em{i.e.,}} BLS, ELM, NeuroFBLS, IF-TSVM, and H-ELM on UCI datasets.}
\label{tab:std_uci}
\resizebox{\textwidth}{!}{%
\begin{tabular}{lccccccc}  \hline \vspace{-2mm}\\ 
\textbf{Dataset $\downarrow$ $\mid$ \text{Model} $\rightarrow$} &
  \textbf{BLS \cite{chen2017broad}} &
  \textbf{ELM \cite{huang2006extreme}} &
  \textbf{NeuroFBLS \cite{feng2018fuzzy}} &
  \textbf{IF-TSVM \cite{rezvani2019intuitionistic}} &
  \textbf{H-ELM \cite{tang2015extreme}} &
  \textbf{F-BLS $^{\dagger}$} &
  \textbf{IF-BLS $^{\dagger}$} \vspace{1mm}\\ \hline \vspace{-2mm}\\
acute\_inflammation               & 0       & 0       & 0       & 6.8465  & 0       & 0       & 0       \\
acute\_nephritis                  & 0       & 0       & 0       & 49.3535 & 20.9165 & 0       & 0       \\
bank                              & 0.6268  & 0.5757  & 0.7048  & 0.5647  & 0.5977  & 0.6556  & 0.6945  \\
breast\_cancer                    & 24.9547 & 34.1162 & 44.6249 & 44.6249 & 20.7893 & 27.5906 & 26.8828 \\
breast\_cancer\_wisc              & 3.9756  & 7.7412  & 5.2416  & 9.6852  & 6.6217  & 5.627   & 5.3499  \\
chess\_krvkp                      & 2.4087  & 8.7903  & 3.6413  & 20.5579 & 12.3882 & 2.6204  & 2.4254  \\
conn\_bench\_sonar\_mines\_rocks & 13.9054      & 5.6916       & 10.5182            & 25.3191          & 34.4077        & 11.3172       & 13.5518        \\
credit\_approval                  & 11.1132 & 10.506  & 10.5832 & 29.4322 & 11.9774 & 11.125  & 10.3423 \\
cylinder\_bands                   & 3.4408  & 3.664   & 6.8956  & 17.9483 & 10.7841 & 3.2817  & 1.5525  \\
echocardiogram                    & 7.467   & 5.8108  & 3.6463  & 7.013   & 6.1721  & 7.2724  & 6.1265  \\
fertility                         & 10      & 9.6177  & 6.7082  & 10.3682 & 8.9443  & 8.2158  & 8.2158  \\
haberman\_survival                & 7.3768  & 8.4751  & 8.4751  & 8.4751  & 8.0641  & 10.0681 & 8.5088  \\
hepatitis                         & 9.8374  & 9.5148  & 5.7705  & 14.1715 & 11.7642 & 9.5148  & 4.2059  \\
hill\_valley                      & 6.8334  & 2.3404  & 4.4314  & 7.2928  & 8.8539  & 3.4346  & 4.279   \\
horse\_colic                      & 4.9082  & 2.5627  & 2.9113  & 5.0495  & 1.4285  & 6.0478  & 2.5582  \\
mammographic                      & 2.4859  & 4.0207  & 5.5859  & 2.7999  & 2.928   & 4.8822  & 5.3901  \\
molec\_biol\_promoter             & 4.3409  & 6.4639  & 8.2564  & 26.2121 & 19.6765 & 3.7913  & 9.5415  \\
monks\_1                          & 11.7835 & 8.4301  & 10.7869 & 5.6022  & 14.7166 & 7.6452  & 9.3321  \\
musk\_1                           & 11.8581 & 4.8493  & 8.7957  & 36.0594 & 6.2829  & 7.25    & 7.7547  \\
oocytes\_merluccius\_nucleus\_4d  & 2.5626  & 2.8977  & 3.7426  & 3.9544  & 4.1503  & 1.5154  & 2.2095  \\
oocytes\_trisopterus\_nucleus\_2f & 4.6837  & 3.9279  & 2.6206  & 8.9408  & 3.7857  & 3.7965  & 2.3332  \\
pima                              & 2.8966  & 2.2956  & 4.2047  & 5.9529  & 4.9923  & 3.8226  & 1.5777  \\
pittsburg\_bridges\_T\_OR\_D      & 5.4205  & 11.5696 & 7.9461  & 13.9488 & 6.524   & 8.3581  & 5.9856  \\
spambase                          & 4.0884  & 5.5405  & 7.5672  & 53.9495 & 9.6543  & 2.8438  & 2.4424  \\
spect                             & 8.3745  & 3.5299  & 6.9325  & 14.0563 & 6.6172  & 2.7986  & 7.3802  \\
statlog\_heart                    & 2.1114  & 3.5621  & 1.0143  & 2.6189  & 3.8401  & 2.8085  & 4.829   \\
tic\_tac\_toe                     & 1.2836  & 7.9105  & 13.1336 & 48.4173 & 25.876  & 1.7038  & 0.8768  \\
titanic                           & 15.5828 & 15.5828 & 13.6955 & 15.9056 & 15.75   & 15.5828 & 15.0381 \vspace{1mm}\\ \hline \vspace{-2mm}\\
Average                           & 6.5829  & 6.7853  & 7.4441  & 17.6829 & 10.3037 & 6.1989  & 6.0494 \vspace{1mm}\\ \hline \vspace{-2mm}\\
\end{tabular}%
}
\end{table*}
\begin{table*}[htp]
\centering
\caption{The rank of the proposed F-BLS and IF-BLS models along with the existing models, {\em{i.e.,}} BLS, ELM, NeuroFBLS, IF-TSVM, and H-ELM on UCI datasets.}
\label{tab:rank_uci}
\resizebox{16cm}{!}{%
\begin{tabular}{lccccccc} \hline \vspace{-2mm}\\ 
        \textbf{Dataset $\downarrow$ $\mid$ \text{Model} $\rightarrow$} &
  \textbf{BLS \cite{chen2017broad}} &
  \textbf{ELM \cite{huang2006extreme}} &
  \textbf{NeuroFBLS \cite{feng2018fuzzy}} &
  \textbf{IF-TSVM \cite{rezvani2019intuitionistic}} &
  \textbf{H-ELM \cite{tang2015extreme}} &
  \textbf{F-BLS $^{\dagger}$} &
  \textbf{IF-BLS $^{\dagger}$} \vspace{1mm}\\ \hline \vspace{-2mm}\\
acute\_inflammation               & 3.5 & 3.5 & 3.5 & 7   & 3.5 & 3.5 & 3.5 \\
acute\_nephritis                  & 3   & 3   & 3   & 7   & 6   & 3   & 3   \\
bank                              & 2   & 4.5 & 3   & 7   & 6   & 1   & 4.5 \\
breast\_cancer                    & 6   & 7   & 4.5 & 4.5 & 1   & 3   & 2   \\
breast\_cancer\_wisc              & 3   & 5   & 1   & 7   & 6   & 4   & 2   \\
chess\_krvkp                      & 2   & 5   & 6   & 7   & 4   & 3   & 1   \\
conn\_bench\_sonar\_mines\_rocks  & 3   & 6   & 5   & 7   & 2   & 4   & 1   \\
credit\_approval                  & 2   & 5   & 6   & 7   & 3   & 4   & 1   \\
cylinder\_bands                   & 2   & 5   & 3   & 7   & 6   & 4   & 1   \\
echocardiogram                    & 4   & 5   & 6   & 7   & 2.5 & 2.5 & 1   \\
fertility                         & 4   & 5.5 & 1   & 7   & 5.5 & 2.5 & 2.5 \\
haberman\_survival                & 6   & 4   & 4   & 4   & 2   & 7   & 1   \\
hepatitis                         & 3   & 5   & 1.5 & 7   & 6   & 4   & 1.5 \\
hill\_valley                      & 1   & 5   & 4   & 7   & 6   & 2   & 3   \\
horse\_colic                      & 3   & 5   & 6   & 7   & 2   & 4   & 1   \\
mammographic                      & 5   & 3   & 2   & 7   & 6   & 4   & 1   \\
molec\_biol\_promoter             & 3   & 5   & 6   & 7   & 4   & 2   & 1   \\
monks\_1                          & 5   & 2   & 1   & 7   & 6   & 4   & 3   \\
musk\_1                           & 4   & 5   & 6   & 7   & 1   & 3   & 2   \\
oocytes\_merluccius\_nucleus\_4d  & 1   & 5   & 4   & 7   & 6   & 2   & 3   \\
oocytes\_trisopterus\_nucleus\_2f & 1   & 3   & 4   & 7   & 6   & 2   & 5   \\
pima                              & 3   & 4   & 2   & 7   & 6   & 5   & 1   \\
pittsburg\_bridges\_T\_OR\_D      & 3   & 5   & 2   & 7   & 6   & 4   & 1   \\
spambase                          & 4   & 5   & 6   & 7   & 2   & 3   & 1   \\
spect                             & 2.5 & 4.5 & 4.5 & 7   & 6   & 2.5 & 1   \\
statlog\_heart                    & 2.5 & 6   & 4.5 & 7   & 4.5 & 2.5 & 1   \\
tic\_tac\_toe                     & 1   & 4   & 5   & 7   & 6   & 2   & 3   \\
titanic                           & 4   & 4   & 2   & 7   & 6   & 4   & 1   \vspace{1mm}\\ \hline \vspace{-2mm}\\
Average & 3.0893       & 4.6071       & 3.8036             & 6.8036           & 4.5357         & 3.2679        & 1.8929       \vspace{1mm}\\ \hline \vspace{-2mm}\\ 
\end{tabular}
}
\end{table*}
\begin{table*}[htp]
\centering
\caption{The best hyperparameters of the proposed F-BLS and IF-BLS models along with the existing models, {\em{i.e.,}} BLS, ELM, NeuroFBLS, IF-TSVM, and H-ELM on UCI datasets.}
\label{tab:PARAMETERS_UCI}
\resizebox{\textwidth}{!}{%
\begin{tabular}{lccccccc}
\hline \vspace{-2mm}\\ 
        \textbf{Dataset $\downarrow$ $\mid$ \text{Model} $\rightarrow$} &
  \textbf{BLS \cite{chen2017broad}} &
  \textbf{ELM \cite{huang2006extreme}} &
  \textbf{NeuroFBLS \cite{feng2018fuzzy}} &
  \textbf{IF-TSVM \cite{rezvani2019intuitionistic}} &
  \textbf{H-ELM \cite{tang2015extreme}} &
  \textbf{F-BLS $^{\dagger}$} &
  \textbf{IF-BLS $^{\dagger}$} \vspace{1mm}\\ \hline \vspace{-2mm}\\ 
 Parameters & ($C, p, m, q$)       & ($C, h_l$)         & ($C, N_{fn}, N_{fg}, q$) & ($C_1, C_2, \mu$) & ($C, h_l$) & ($C, p, m, q$)            & ($C, \mu, p, m, q$)         \vspace{1mm}\\ \hline \vspace{-2mm}\\ 
acute\_inflammation &
  ($0,5,9,45$) &
  ($0.000001,15$) &
  ($0.000001,5,3,5$) &
  ($0.000001,10000,0.03125$) &
  ($0.01,25$) &
  ($0.000001,5,5,35$) &
  ($100,0.03125,5,3,85$) \\
acute\_nephritis &
  ($1,5,5,35$) &
  ($1,45$) &
  ($0.000001,5,1,45$) &
  ($0.000001,10000,0.03125$) &
  ($0.0001,85$) &
  ($1,5,3,105$) &
  ($10000,0.03125,15,9,35$) \\
bank &
  ($1000000,25,21,85$) &
  ($10000,195$) &
  ($0.000001,15,15,45$) &
  ($0.000001,1000000,1$) &
  ($0.01,25$) &
  ($1000000,15,15,65$) &
  ($1000000,0.5,45,19,15$) \\
breast\_cancer &
  ($0,15,3,105$) &
  ($0.000001,15$) &
  ($100,5,1,5$) &
  ($0.000001,0.000001,0.03125$) &
  ($0.000001,35$) &
  ($0.000001,40,1,55$) &
  ($1000000,8,5,3,75$) \\
breast\_cancer\_wisc &
  ($1,25,15,85$) &
  ($1,155$) &
  ($1,30,17,55$) &
  ($0.000001,10000,0.03125$) &
  ($1,45$) &
  ($1,35,13,75$) &
  ($1000000,1,20,21,85$) \\
chess\_krvkp &
  ($0.01,20,3,45$) &
  ($10000,165$) &
  ($0.000001,40,21,15$) &
  ($1000000,0.000001,0.03125$) &
  ($0.000001,75$) &
  ($0.01,40,5,15$) &
  ($1000000,0.03125,30,3,75$) \\
conn\_bench\_sonar\_mines\_rocks &
  ($0.01,5,3,95$) &
  ($10000,165$) &
  ($0.0001,10,17,5$) &
  ($10000,0.000001,0.03125$) &
  ($0.0001,135$) &
  ($0.0001,10,1,95$) &
  ($1,16,50,15,105$) \\
credit\_approval &
  ($0.01,15,1,5$) &
  ($10000,45$) &
  ($1,5,21,15$) &
  ($0.000001,0.000001,0.03125$) &
  ($0.01,35$) &
  ($0.01,5,15,65$) &
  ($100,4,10,1,85$) \\
cylinder\_bands &
  ($1,25,1,85$) &
  ($0.01,125$) &
  ($0.01,15,13,95$) &
  ($0.000001,0.000001,0.03125$) &
  ($0.01,55$) &
  ($1,50,13,85$) &
  ($100,4,5,19,65$) \\
echocardiogram &
  ($1000000,20,1,15$) &
  ($0.01,155$) &
  ($1,15,3,105$) &
  ($0.000001,1000000,0.03125$) &
  ($0.01,125$) &
  ($100,10,1,105$) &
  ($1000000,8,50,19,5$) \\
fertility &
  ($100,10,1,75$) &
  ($0.000001,55$) &
  ($1,10,9,35$) &
  ($0.000001,0.000001,0.03125$) &
  ($0.01,115$) &
  ($100,45,1,85$) &
  ($1000000,1,40,17,85$) \\
haberman\_survival &
  ($1,25,21,95$) &
  ($0.000001,5$) &
  ($1,5,1,5$) &
  ($0.000001,0.000001,0.03125$) &
  ($0.0001,55$) &
  ($1000000,45,3,5$) &
  ($1000000,16,35,1,75$) \\
hepatitis &
  ($100,5,3,75$) &
  ($0.01,135$) &
  ($0.0001,45,17,55$) &
  ($0.000001,0.000001,0.03125$) &
  ($0.01,105$) &
  ($1,40,19,35$) &
  ($10000,8,30,5,75$) \\
hill\_valley &
  ($0.01,35,9,5$) &
  ($0.01,205$) &
  ($1,45,21,25$) &
  ($10000,1000000,0.125$) &
  ($0.01,195$) &
  ($1,25,19,75$) &
  ($1,32,50,5,55$) \\
horse\_colic &
  ($1,35,1,55$) &
  ($0.01,175$) &
  ($1,5,19,95$) &
  ($0.000001,0.000001,0.03125$) &
  ($1,145$) &
  ($1,35,7,85$) &
  ($1000000,2,50,17,15$) \\
mammographic &
  ($10000,10,21,45$) &
  ($1000000,75$) &
  ($1,15,13,15$) &
  ($0.000001,0.000001,0.03125$) &
  ($0.0001,5$) &
  ($10000,30,1,105$) &
  ($1000000,0.25,10,19,85$) \\
molec\_biol\_promoter &
  ($0.01,15,15,105$) &
  ($0.01,205$) &
  ($0.0001,25,3,45$) &
  ($0.000001,100,0.03125$) &
  ($0.0001,35$) &
  ($1,20,9,5$) &
  ($100,8,30,7,15$) \\
monks\_1 &
  ($1000000,10,1,95$) &
  ($100,205$) &
  ($1,50,9,75$) &
  ($0.000001,0.000001,0.03125$) &
  ($0.01,135$) &
  ($1000000,10,1,95$) &
  ($1000000,0.5,45,19,85$) \\
musk\_1 &
  ($0.01,10,3,25$) &
  ($0.0001,205$) &
  ($0.01,35,21,5$) &
  ($10000,0.000001,0.03125$) &
  ($0.0001,175$) &
  ($0.01,30,13,5$) &
  ($10000,16,10,17,65$) \\
oocytes\_merluccius\_nucleus\_4d &
  ($1,35,19,55$) &
  ($0.01,175$) &
  ($0.0001,5,21,45$) &
  ($1000000,0.000001,0.03125$) &
  ($1,165$) &
  ($1,30,21,65$) &
  ($1000000,0.0625,40,21,55$) \\
oocytes\_trisopterus\_nucleus\_2f &
  ($1000000,30,1,25$) &
  ($0.01,205$) &
  ($0.0001,5,19,15$) &
  ($1000000,0.000001,0.03125$) &
  ($0.01,145$) &
  ($100,40,1,75$) &
  ($1000000,0.5,45,19,45$) \\
pima &
  ($100,30,1,25$) &
  ($1000000,105$) &
  ($0.01,5,9,65$) &
  ($0.000001,0.000001,0.03125$) &
  ($1,55$) &
  ($100,20,1,5$) &
  ($1000000,1,10,9,95$) \\
pittsburg\_bridges\_T\_OR\_D &
  ($1,30,21,65$) &
  ($0.0001,15$) &
  ($1,40,1,75$) &
  ($0.000001,0.000001,0.03125$) &
  ($0.0001,125$) &
  ($100,30,3,15$) &
  ($10000,2,40,5,95$) \\
spambase &
  ($0.01,30,19,65$) &
  ($100,175$) &
  ($0.000001,5,19,75$) &
  ($0.000001,0.000001,0.03125$) &
  ($0.01,75$) &
  ($0.01,10,9,55$) &
  ($1,16,45,19,15$) \\
spect &
  ($0.01,5,13,55$) &
  ($0.000001,85$) &
  ($0.01,45,11,55$) &
  ($0.000001,0.000001,0.03125$) &
  ($1,35$) &
  ($1,35,5,95$) &
  ($100,16,25,9,25$) \\
statlog\_heart &
  ($1,30,9,105$) &
  ($0.01,55$) &
  ($1,40,15,105$) &
  ($0.000001,0.000001,0.03125$) &
  ($0.01,55$) &
  ($1,50,7,105$) &
  ($10000,4,25,17,55$) \\
tic\_tac\_toe &
  ($1,15,21,105$) &
  ($10000,205$) &
  ($0.01,35,15,95$) &
  ($0.000001,0.000001,0.03125$) &
  ($0.000001,85$) &
  ($1,15,21,45$) &
  ($10000,2,10,21,55$) \\
titanic &
  ($10000,10,3,65$) &
  ($0.000001,75$) &
  ($100,5,1,55$) &
  ($0.000001,10000,0.03125$) &
  ($0.01,5$) &
  ($0.000001,5,1,5$) &
  ($1,2,15,1,65$)  \vspace{1mm}\\ \hline \vspace{-2mm}\\ 
\end{tabular}%
}
\end{table*}
\begin{table*}[htp]
\centering
\caption{The standard deviations of the proposed F-BLS and IF-BLS models along with the existing models, {\em{i.e.,}} BLS, ELM, NeuroFBLS, IF-TSVM, and H-ELM on UCI dataset with varying levels of $5\%$, $10\%$, $15\%$, and $20\%$ Gaussian noise.}
\label{tab:Noise_std}
\resizebox{\textwidth}{!}{%
\begin{tabular}{lcccccccc}\hline \vspace{-2mm}\\ 
\textbf{Dataset $\downarrow$ $\mid$ \text{Model} $\rightarrow$} &
  \textbf{Noise} &
  \textbf{BLS \cite{chen2017broad}} &
  \textbf{ELM \cite{huang2006extreme}} &
  \textbf{NeuroFBLS \cite{feng2018fuzzy}} &
  \textbf{IF-TSVM \cite{rezvani2019intuitionistic}} &
  \textbf{H-ELM \cite{tang2015extreme}} &
  \textbf{F-BLS $^{\dagger}$} &
  \textbf{IF-BLS $^{\dagger}$} \vspace{1mm}\\ \hline \vspace{-2mm}\\
\multirow{4}{*}{breast\_cancer} & $5\%$          & 26.0578      & 43.5886      & 44.6249            & 44.6249          & 21.5082        & 27.9532        & 23.9395         \\
                                 & $10\%$               & 27.547  & 45.6714 & 44.6249 & 44.6249 & 20.109  & 27.6671 & 24.5976 \\
                                 & $15\%$               & 27.0047 & 44.3655 & 43.9368 & 44.6249 & 27.4997 & 24.3874 & 21.4006 \\
                                 & $20\%$               & 26.8794 & 42.3351 & 44.4418 & 44.6249 & 30.929  & 29.7448 & 23.919  \vspace{1mm}\\ \hline \vspace{-2mm}\\
conn\_bench\_sonar\_mines\_rocks & $5\%$                & 9.8007  & 3.9134  & 7.7724  & 21.9294 & 31.0739 & 9.4446  & 8.9616  \\
\textbf{}                        & $10\%$               & 14.8713 & 12.8776 & 10.5432 & 24.0628 & 32.2191 & 7.7664  & 13.866  \\
                                 & $15\%$               & 6.373   & 7.4754  & 13.0538 & 16.9866 & 13.0222 & 6.4177  & 8.7652  \\
                                 & $20\%$               & 12.7508 & 5.4069  & 20.556  & 22.0657 & 26.5379 & 8.8397  & 15.2157 \vspace{1mm}\\ \hline \vspace{-2mm}\\
hill\_valley                     & $5\%$                & 0.4903  & 2.3907  & 3.1142  & 4.2855  & 6.9003  & 0.7357  & 3.366   \\
                                 & $10\%$               & 2.3794  & 3.7976  & 0.4881  & 4.3653  & 4.1687  & 3.1585  & 2.092   \\
                                 & $15\%$               & 1.961   & 2.4486  & 2.9256  & 1.5395  & 7.9389  & 1.9428  & 1.9166  \\
                                 & $20\%$               & 2.419   & 2.1926  & 3.0069  & 2.0268  & 6.4379  & 2.4648  & 2.2579  \vspace{1mm}\\ \hline \vspace{-2mm}\\
pittsburg\_bridges\_T\_OR\_D     & $5\%$                & 9.5855  & 13.9488 & 9.5615  & 13.9488 & 8.6481  & 8.4472  & 4.4237  \\
                                 & $10\%$               & 8.3581  & 10.0785 & 7.9461  & 13.9488 & 13.9488 & 6.4957  & 5.6944  \\
                                 & $15\%$               & 8.9271  & 11.9712 & 8.3734  & 13.9488 & 13.9488 & 4.5094  & 4.4639  \\
                                 & $20\%$               & 7.4357  & 13.9488 & 5.6944  & 13.9488 & 13.4466 & 9.2637  & 5.5185  \vspace{1mm}\\ \hline \vspace{-2mm}\\
tic\_tac\_toe                    & $5\%$                & 2.4968  & 20.0315 & 9.7246  & 48.3136 & 29.0264 & 2.1695  & 1.6115  \\
                                 & $10\%$               & 2.0868  & 11.8785 & 10.5162 & 48.4173 & 24.1614 & 1.172   & 1.6671  \\
                                 & $15\%$               & 2.3264  & 9.2264  & 13.3383 & 48.4173 & 37.2164 & 3.0788  & 1.5491  \\
                                 & $20\%$               & 2.0997  & 14.1505 & 14.1961 & 48.4173 & 43.2215 & 3.362   & 1.9521  \vspace{1mm}\\ \hline \vspace{-2mm}\\
Avg. Std. Dev.                   & \multicolumn{1}{l}{} & 10.0925&	16.0849& 15.92196&	26.2561&	20.5981&	9.4511 & 8.8589 \vspace{1mm}\\ \hline \vspace{-2mm}
\end{tabular}%
}
\end{table*}
\begin{table*}[htp]
\centering
\caption{The hyperparameters of the proposed F-BLS and IF-BLS models along with the existing models, {\em{i.e.,}} BLS, ELM, NeuroFBLS, IF-TSVM, and H-ELM on UCI dataset with varying levels of $5\%$, $10\%$, $15\%$, and $20\%$ Gaussian noise.}
\label{tab:Noise_Hyperparameters}
\resizebox{\textwidth}{!}{%
\begin{tabular}{lcccccccc}\hline \vspace{-2mm}\\ 
\textbf{Dataset $\downarrow$ $\mid$ \text{Model} $\rightarrow$} &
  \textbf{Noise $\downarrow$} &
  \textbf{BLS \cite{chen2017broad}} &
  \textbf{ELM \cite{huang2006extreme}} &
  \textbf{NeuroFBLS \cite{feng2018fuzzy}} &
  \textbf{IF-TSVM \cite{rezvani2019intuitionistic}} &
  \textbf{H-ELM \cite{tang2015extreme}} &
  \textbf{F-BLS $^{\dagger}$} &
  \textbf{IF-BLS $^{\dagger}$} \vspace{1mm}\\ \hline \vspace{-2mm}\\ 
Parameters $\rightarrow$ &  & ($C, p, m, q$)       & ($C, h_l$)         & ($C, N_{fn}, N_{fg}, q$) & ($C_1, C_2, \mu$) & ($C, h_l$) & ($C, p, m, q$)            & ($C, \mu, p, m, q$)         \vspace{1mm}\\ \hline \vspace{-2mm}\\
breast\_cancer &
  $5\%$ &
  ($0.0001,10,3,105$) &
  ($0.000001,5$) &
  ($100,5,1,5$) &
  ($0.000001,0.000001,0.03125$) &
  ($0.000001,125$) &
  ($0.0001,10,1,95$) &
  ($100,16,5,1,25$) \\
 &
  $10\%$ &
  ($0,40,1,25$) &
  ($0.000001,5$) &
  ($100,5,1,5$) &
  ($0.000001,0.000001,0.03125$) &
  ($0.000001,5$) &
  ($0.000001,25,1,15$) &
  ($1000000,8,10,1,5$) \\
 &
  $15\%$ &
  ($0,25,1,35$) &
  ($0.0001,5$) &
  ($100,20,1,25$) &
  ($0.000001,0.000001,0.03125$) &
  ($0.0001,95$) &
  ($0.0001,5,3,105$) &
  ($1000000,4,5,3,85$) \\
 &
  $20\%$ &
  ($0,15,1,65$) &
  ($0.0001,5$) &
  ($100,35,17,35$) &
  ($0.000001,0.000001,0.03125$) &
  ($0.000001,15$) &
  ($0.0001,15,3,5$) &
  ($10000,4,15,3,45$) \vspace{1mm}\\ \hline \vspace{-2mm}\\
conn\_bench\_sonar\_mines\_rocks &
  $5\%$ &
  ($0.01,10,1,95$) &
  ($100,115$) &
  ($0.0001,15,13,75$) &
  ($10000,0.000001,0.03125$) &
  ($0.000001,175$) &
  ($0.01,35,3,35$) &
  ($1000000,32,50,1,105$) \\
 &
  $10\%$ &
  ($0.01,5,9,95$) &
  ($1000000,145$) &
  ($0.000001,10,17,95$) &
  ($10000,0.000001,0.03125$) &
  ($0.000001,155$) &
  ($10000,40,9,15$) &
  ($100,16,35,11,5$) \\
 &
  $15\%$ &
  ($0.01,5,3,55$) &
  ($10000,155$) &
  ($0.0001,20,11,95$) &
  ($10000,0.000001,0.03125$) &
  ($0.0001,115$) &
  ($0.0001,50,1,45$) &
  ($1,32,5,19,15$) \\
 &
  $20\%$ &
  ($0.01,5,17,95$) &
  ($10000,75$) &
  ($0.0001,15,11,25$) &
  ($10000,0.000001,0.03125$) &
  ($0.0001,195$) &
  ($10000,35,21,15$) &
  ($10000,32,50,1,55$) \vspace{1mm}\\ \hline \vspace{-2mm}\\
hill\_valley &
  $5\%$ &
  ($0.01,20,17,55$) &
  ($0.01,175$) &
  ($0.000001,10,17,85$) &
  ($100,100,16$) &
  ($0.01,195$) &
  ($1,35,9,85$) &
  ($1000000,0.0625,25,21,75$) \\
 &
  $10\%$ &
  ($0.01,50,19,55$) &
  ($0.01,205$) &
  ($1,15,19,15$) &
  ($10000,1000000,0.03125$) &
  ($0.01,155$) &
  ($0.01,25,17,75$) &
  ($1000000,0.03125,20,19,25$) \\
 &
  $15\%$ &
  ($0.01,30,19,25$) &
  ($0.01,205$) &
  ($1,35,17,95$) &
  ($10000,1000000,0.03125$) &
  ($0.01,135$) &
  ($0.01,30,11,105$) &
  ($1000000,0.03125,50,19,95$) \\
 &
  $20\%$ &
  ($0.01,45,19,95$) &
  ($0.01,195$) &
  ($1,35,21,45$) &
  ($0.000001,1000000,0.03125$) &
  ($0.01,145$) &
  ($0.01,40,21,55$) &
  ($1000000,0.03125,20,15,25$) \vspace{1mm}\\ \hline \vspace{-2mm}\\
pittsburg\_bridges\_T\_OR\_D &
  $5\%$ &
  ($100,40,1,55$) &
  ($0.000001,5$) &
  ($1,30,5,55$) &
  ($0.000001,0.000001,0.03125$) &
  ($0.01,165$) &
  ($100,15,3,105$) &
  ($100,32,20,1,85$) \\
 &
  $10\%$ &
  ($1000000,45,1,25$) &
  ($1,115$) &
  ($1,50,1,15$) &
  ($0.000001,0.000001,0.03125$) &
  ($0.000001,5$) &
  ($100,30,9,85$) &
  ($100,32,40,3,85$) \\
 &
  $15\%$ &
  ($100,5,3,65$) &
  ($1000000,35$) &
  ($0.01,15,1,5$) &
  ($0.000001,0.000001,0.03125$) &
  ($0.000001,5$) &
  ($100,20,3,45$) &
  ($10000,2,45,11,15$) \\
 &
  $20\%$ &
  ($100,45,1,25$) &
  ($0.000001,5$) &
  ($1,40,7,15$) &
  ($0.000001,0.000001,0.03125$) &
  ($0.01,145$) &
  ($10000,30,21,45$) &
  ($10000,2,30,5,105$) \vspace{1mm}\\ \hline \vspace{-2mm}\\
tic\_tac\_toe &
  $5\%$ &
  ($1,10,21,105$) &
  ($1,175$) &
  ($0.000001,15,19,25$) &
  ($1000000,0.000001,0.03125$) &
  ($0.0001,25$) &
  ($100,10,19,75$) &
  ($1000000,1,30,21,105$) \\
 &
  $10\%$ &
  ($1,15,21,105$) &
  ($1000000,185$) &
  ($0.000001,15,19,35$) &
  ($0.000001,0.000001,0.03125$) &
  ($0.000001,195$) &
  ($100,35,19,105$) &
  ($1000000,1,40,21,105$) \\
 &
  $15\%$ &
  ($1,25,21,95$) &
  ($100,195$) &
  ($0.01,25,13,65$) &
  ($0.000001,0.000001,0.03125$) &
  ($0.0001,155$) &
  ($1,25,21,85$) &
  ($1000000,1,25,21,75$) \\
 &
  $20\%$ &
  ($1,25,21,105$) &
  ($1,185$) &
  ($0.000001,20,15,55$) &
  ($0.000001,0.000001,0.03125$) &
  ($0.0001,105$) &
  ($100,10,19,75$) &
  ($1000000,1,10,21,75$) \vspace{1mm}\\ \hline \vspace{-2mm}
\end{tabular}%
}
\end{table*}
\begin{table*}[htp]
\centering
\caption{The best hyperparameters of the proposed F-BLS and IF-BLS models along with the existing models, {\em{i.e.,}} BLS, ELM, NeuroFBLS, IF-TSVM, and H-ELM on ADNI datasets.}
\label{tab:AD_Parameters}
\resizebox{\textwidth}{!}{%
\begin{tabular}{lccccccc}
          \hline \vspace{-2mm}\\ 
        \textbf{Dataset $\downarrow$ $\mid$ \text{Model} $\rightarrow$} &
  \textbf{BLS \cite{chen2017broad}} &
  \textbf{ELM \cite{huang2006extreme}} &
  \textbf{NeuroFBLS \cite{feng2018fuzzy}} &
  \textbf{IF-TSVM \cite{rezvani2019intuitionistic}} &
  \textbf{H-ELM \cite{tang2015extreme}} &
  \textbf{F-BLS $^{\dagger}$} &
  \textbf{IF-BLS $^{\dagger}$} \vspace{1mm}\\ \hline \vspace{-2mm}\\ 
 Parameters & ($C, p, m, q$)       & ($C, h_l$)         & ($C, N_{fn}, N_{fg}, q$) & ($C_1, C_2, \mu$) & ($C, h_l$) & ($C, p, m, q$)            & ($C, \mu, p, m, q$)         \vspace{1mm}\\ \hline \vspace{-2mm}\\ 
CN\_VS\_AD  & ($1,30,11,35$) & ($100,95$)   & ($0.01,10,9,85$)   & ($0.000001,1000000,0.03125$)  & ($1,105$)      & ($10000,35,3,55$) & ($100,2,15,19,95$) \\
CN\_VS\_MCI & ($100,35,19,55$) & ($10000,105$) & ($0.0001,30,3,95$) & ($0.000001,0.000001,0.03125$) & ($0.01,115$) & ($100,45,17,105$) & ($10000,4,30,7,15$) \\
MCI\_VS\_AD & ($1,40,9,45$)  & ($1,125$)    & ($1,10,5,85$)      & ($0.000001,0.000001,0.03125$) & ($0.01,145$)   & ($1,35,21,45$)    & ($100,16,30,3,65$) \vspace{1mm}\\ \hline \vspace{-2mm}\\ 
\end{tabular}%
}
\end{table*}
\clearpage
\newpage
\bibliographystyle{IEEEtranN}
 \bibliography{refs}